%% file: ms.tex
\newif\ifarxiv
\arxivtrue

\ifarxiv
\documentclass{article}
\usepackage{arxiv}
\usepackage{appendix}
\else
\documentclass[journal]{IEEEtran}
\fi

\usepackage[T1]{fontenc}

\input{latex-math/basic-math}
\input{latex-math/basic-ml}
\input{latex-math/ml-automl}

\input{latex-math/ml-mbo}

\usepackage{fancyvrb}
\usepackage{booktabs}
\usepackage{pifont}

\usepackage{cite}

\ifarxiv
\iftrue
\else
\ifCLASSINFOpdf
  \usepackage[pdftex]{graphicx}
\else
\fi
\fi

\usepackage{amsmath}
\usepackage{dsfont}
\usepackage{amssymb}
\usepackage{bm}

\usepackage{algorithm}
\usepackage{algpseudocode}

\usepackage{array}

\usepackage{subcaption}

\usepackage{url}

\usepackage{amsmath}

\usepackage{todonotes}
\usepackage{multirow}
\usepackage{lscape}

\usepackage[inline,shortlabels]{enumitem}
\usepackage[subtle]{savetrees}
\begin{document}
\title{Automated Benchmark-Driven Design and Explanation of Hyperparameter Optimizers}
%

\author{Julia~Moosbauer*$^1$,
        Martin~Binder*$^1$,  \\
        Lennart~Schneider$^1$, 
        Florian~Pfisterer$^1$,
        Marc~Becker$^1$,
        Michel~Lang$^1$,
        Lars~Kotthoff$^2$,
        Bernd~Bischl$^1$%
\ifarxiv
\\
\\
$^1$ Department of Statistics\\
Ludwig-Maximilians-Universität München\\
\texttt{[first.last]@stat.uni-muenchen.de}\\
\\
$^2$ Department of Computer Science\\
University of Wyoming\\
\texttt{larsko@uwyo.edu}
\else\thanks{*Equal Contribution}\fi
}

\markboth{IEEE TEVC Special Issue: Benchmarking Sampling-Based Optimization Heuristics}%
{Shell \MakeLowercase{\textit{et al.}}: Bare Demo of IEEEtran.cls for IEEE Journals}
%



\maketitle
\ifarxiv
\renewcommand*{\thefootnote}{*}
\footnotetext{These authors contributed equally to this work.}
\renewcommand*{\thefootnote}{\arabic{footnote}}
\fi
\begin{abstract}
\input{00_abstract}
\end{abstract}

\ifarxiv
\else
\begin{IEEEkeywords}
Algorithm design, algorithm analysis, hyperparameter optimization, multifidelity, automated machine learning
\end{IEEEkeywords}

%
\IEEEpeerreviewmaketitle
\fi
\newcommand{\todocite}[1]{}

\section{Introduction}
\label{sec:10_introduction}
\input{10_introduction}

\section{Related Work}
\label{sec:20_related_work}
\input{20_related_work}

\section{HPO and Multifidelity HPO}
\label{sec:25_methodology}
\input{25_methodology}

\section{Formalizing 
a broad class of MF-HPO ALgorithms}
\label{sec:30_multifid_hpo}
\input{30_multifid_hpo}

\section{Automatic Algorithm Design and Analysis}
\label{sec:35_design_and_analysis}
\input{35_design_and_analysis}

\section{Experimental Setup}
\label{sec:40_experimental_design}
\input{40_experimental_design}

\section{Results}
\label{sec:50_results}
\input{50_results}

\section{Conclusion}
\label{sec:90_conclusion}
\input{90_conclusion}
\section*{Acknowledgment}
The authors of this work take full responsibilities for its content.
This work was supported by the German Federal Ministry of Education and Research (BMBF) under Grant No. 01IS18036A. LK is supported by NSF grant 1813537.

\clearpage


\bibliographystyle{IEEEtran}
\bibliography{main}

\clearpage

\appendices
\input{95_appendix}



\ifarxiv
\iftrue
\else
\ifCLASSOPTIONcaptionsoff
  \newpage
\fi
\fi


\end{document}

%% file: latex-math/basic-math.tex
\ifdefined\N                                                                
\renewcommand{\N}{\mathds{N}}                                                
\else
  \newcommand{\N}{\mathds{N}}
\fi
\newcommand{\R}{\mathds{R}}                                                 
\ifdefined\C
  \renewcommand{\C}{\mathds{C}}                                             
\else
  \newcommand{\C}{\mathds{C}}
\fi

\def\argmin{\mathop{\sf arg\,min}}                                          

\newcommand{\xv}{\mathbf{x}}													

\renewcommand{\P}{\mathds{P}}                                               
\newcommand{\E}{\mathds{E}}                                                 

%% file: latex-math/basic-ml.tex
\newcommand{\Xspace}{\mathcal{X}}                                           
\newcommand{\Yspace}{\mathcal{Y}}                                           
\newcommand{\Pxy}{\P_{xy}}                                                  
\newcommand{\D}{\mathcal{D}}                                                
\renewcommand{\xi}[1][i]{\mathbf{x}^{(#1)}}                                          
\newcommand{\yi}[1][i]{y^{(#1)}}                                            

\newcommand{\inducer}{\ensuremath{\mathcal{I}}}                                                

\newcommand{\fh}{\hat{f}}                                                   

%% file: latex-math/ml-automl.tex
\newcommand{\lambdav}{\bm{\lambda}}											

%% file: latex-math/ml-mbo.tex
\newcommand{\archive}{\mathcal{A}}                          

%% file: 00_abstract.tex
Automated hyperparameter optimization (HPO) has gained great popularity and is an important ingredient of most automated machine learning frameworks.

The process of designing HPO algorithms, however, is still an unsystematic and manual process: Limitations of prior work are identified and the improvements proposed are -- even though guided by expert knowledge -- still somewhat arbitrary. This rarely allows for gaining a holistic understanding of which algorithmic components are driving performance, and carries the risk of overlooking good algorithmic design choices. 

We present a principled approach to automated benchmark-driven algorithm design applied to multifidelity HPO (MF-HPO): First, we formalize a rich space of MF-HPO candidates that includes, but is not limited to common HPO algorithms, and then present a configurable framework covering this space. To find the best candidate automatically and systematically, we follow a programming-by-optimization approach and search over the space of algorithm candidates via Bayesian optimization. We challenge whether the found design choices are necessary or could be replaced by more naive and simpler ones by performing an ablation analysis. 
We observe that using a relatively simple configuration, in some ways simpler than established methods, performs very well as long as some critical configuration parameters have the right value.

%% file: 10_introduction.tex
Machine learning (ML) is, in many regards, an optimization problem, and many ML methods can be expressed as algorithms that perform loss minimization with respect to a given objective function. The higher-level task of selecting the ML method and its configuration is often framed as an optimization problem as well, sometimes referred to simply as \emph{hyperparameter optimization} (HPO)~\cite{bischl21hpo} or \emph{combined algorithm selection and hyperparameter optimization} (CASH) problem~\cite{kotthoff2019autoweka}. 
Successfully addressing this problem can lead to large performance gains compared to simply using a default set of parameters. Automated methods to perform this meta-optimization, in the form of automated machine learning (AutoML), can make ML methods more accessible to non-experts.
Because of their potential benefits to ML performance and usability, it is of particular interest to find optimization algorithms that perform well for this meta-optimization. 

Optimization problems arise in many fields of science and engineering, but as the no-free-lunch theorem states, there is no one optimization algorithm that solves all problems equally well~\cite{wolpert_no_1997}. To design suitable optimizers, it is therefore important to understand the characteristics of HPO:
\begin{itemize}
    \item \textbf{Black-box}: The objective usually provides no analytical information~\cite{jones98ego} -- such as a gradient -- thus rendering the application of many traditional optimization methods, such as BFGS, inappropriate or at least dubious.
    \item \textbf{Complex Search Space}: The search space of the optimization problem is often high-dimensional and may contain continuous, integer-valued and categorical dimensions.
    Often, there are dependencies between dimensions or even specific parameter values~\cite{hutter11smac}. 
    \item \textbf{Expensive}: A single evaluation of the objective function may take hours or days. Thus, the total number of possible function evaluations is often severely limited~\cite{jones98ego}. 
    \item \textbf{Low-fidelity approximations possible}: An approximation of the true objective value at lower expense can often be obtained, for example through a partial evaluation~\cite{swersky14freezethaw}.
    \item \textbf{Low effective dimensionality}: The landscape of the objective function can usually be approximated with high fidelity as a function of a small subset of all dimensions~\cite{bergstra2012randomsearch}.
\end{itemize}

These characteristics have been a primary focus in recent HPO and AutoML research.
Many approaches tackle HPO by estimating a local or global structure of the objective landscape by some form of statistical or ML model. This introduces additional overhead and increases the overall resource requirements of the method compared to e.g.\ random search.

Bayesian optimization (BO)~\cite{snoek12practicalbo} and frameworks based on BO are usually based on the global optimization of a non-linear regression model, e.g., a Gaussian process or random forest.
They have yielded significant improvements in performance~\cite{turner20bovsrs} by putting less emphasis on unpromising parts of the search space, but carry a significant overhead. 
BO is somewhat difficult to parallelize due to its sequential nature, although many variants exists e.g.~\cite{hutter_parallel_2012,bischl14moimbo,gonzalez16batchbo,chevalier_fast_2013,balandat2020botorch}, for which superiority is unclear.


\emph{Multifidelity} HPO (MF-HPO) algorithms aim to accelerate the optimization process by exploiting cheaper proxy functions of the objective function itself (e.g., by training ML models on a smaller subsample of the available training data, or by running fewer training iterations). Bandit-based algorithms like Hyperband (HB)~\cite{li2017hyperband} have become particularly popular because of their good trade-off between optimization performance and simplicity.
In more recent works, the expensive nature of the optimization problem is addressed by algorithms designed with both parallel execution and multifidelity evaluations~\cite{li2017hyperband, li20asha, klein20abohb}

Progress in the field of MF-HPO often consists of iterative improvements of established algorithms. Considerable work has been done, for example, to improve the limitations of HB: ASHA~\cite{li20asha} proposes a sophisticated way to make efficient use of parallel resources, BOHB~\cite{falkner18bohb} improves performance during later parts of a run by incorporating surrogate assistance into HB, and A-BOHB \cite{tiao20abohb} unites a bandit-based optimization scheme using model-based guidance with asynchronous parallelization.

While these conceptual extensions of HPO carry all their respective merit, it is often somewhat overlooked that the simplicity of an optimization algorithm heavily influences its adoption in practice, i.e.\ how difficult modifications and extensions are and on how many dependencies a system relies~\cite{sculley15technicaldept}. 
For example, random search (RS) still enjoys great popularity, as it is extremely simple to implement and parallelize, has almost no overhead, and is able to take advantage of the aforementioned low-effective dimensionality~\cite{bergstra2012randomsearch}.
Furthermore, algorithmic developments build on prior research that identifies and addresses limitations, but rarely questions core algorithmic choices that have been made in the original implementation. Many multifidelity algorithms, for example, are extensions and further developments of HB that take the fixed successive halving schedule for granted. 
The process of designing a good MF-HPO optimizer in practice, and many other algorithmic solutions in science in general, can therefore often feel somewhat like a ``manual stochastic local search on the meta level''. 
The drawback of this manual process is that the design space of all HPO algorithms is not searched systematically and parts of it excluded by prior algorithmic decisions. This carries the risk that well-performing algorithms are overlooked, as ``established'' algorithmic choices are not questioned, and it is often hard to determine which components of an algorithm make a difference. In particular, it is possible that overly complicated algorithms with many components are developed by extending ``established'' designs, only some of which contribute meaningfully to performance gains. We design an HPO algorithm following the programming-by-optimization paradigm~\cite{hoos_programming_2012}. We formalize the space of potential HPO algorithms to be able to apply an automatic search procedure to find the best designs of optimization algorithm. Furthermore, as any HPO algorithm is going to be applied in a diverse set of application scenarios that are very different (e.g.\ in terms of dimensionality and complexity of the search space), we require an expressive and representative benchmark on which to perform the optimization and evaluation. 

\subsection{Contributions}

We demonstrate in a principled manner how HPO algorithm design can be performed systematically and automatically in a benchmark-driven approach. In particular, the contributions of this work are: 

\begin{itemize}
    \item \textbf{Formalization}: We formalize the design space of MF-HPO algorithms and demonstrate how common MF-HPO algorithms represent instances within this space.

    \item \textbf{Framework}: Based on this formalization, we present a rich, configurable framework for MF-HPO algorithms, whose software implementation is called \textsc{Smashy} (Surrogate Model Assisted HYperband).
    \item \textbf{Configuration}: Based on the formalization and framework, we follow an empirical approach to design an MF-HPO algorithm by optimization, given a 2 large benchmarks suits. This configuration procedure does not only consider performance, but also e.g.\ the simplicity of the design.
    \item \textbf{Explanation}: For the resulting MF-HPO system, we systematically assess and explain the effect of different design choices on overall algorithmic performance.
    Furthermore, We investigate the behavior of algorithmic design components in the context of specific problem scenarios; i.e., we investigate which algorithmic components lead to performance improvements for simple HPO with numeric hyperparameters, AutoML pipeline configuration, and neural architecture search. 
\end{itemize}


We stress that the design of a new algorithm for algorithm configuration is not in the scope of this paper. We are building on prior work and use state-of-the-art configuration algorithms, and apply them to our novel MF-HPO framework.

%% file: 20_related_work.tex


AutoML is a relatively new field, enabled by recent
advances in computational resources, the availability of data and datasets for
ML on a large scale, and driven by an ever-growing need for
ML expertise that cannot be met by human experts. However, the
basic problem of choosing the most suitable approach for solving a problem goes
back much further than that, formalized as the Algorithm Selection Problem in
its most basic form~\cite{rice_algorithm_1976}.
Before the advent of large-scale computing and data, so-called meta-learning
methods were developed to make the best use of the sparse experience of running
ML on specific tasks. There are many approaches; one of the most
popular identifies approaches that performed well on tasks similar to the one
given and applies them (see, e.g., \cite{pfahringer_meta-learning_2000,sun_pairwise_2013,soares_meta-learning_2004}).
There are more recent approaches to
meta-learning~\cite{finn_model-agnostic_2017} for deep neural networks. The
interested reader is referred to a recent survey on
meta-learning~\cite{vanschoren2018metalearning}.
Arguably, one of the most successful and popular tools for current AutoML is HPO~\cite{bischl21hpo} and its recent variant MF-HPO. As we reference both extensively in Section~\ref{sec:10_introduction} and the remainder of the paper deals exclusive with this topic, we abstain from a further discussion here.
Although new, AutoML has become a large field, and a full review is beyond the scope
of this paper, the interested reader is referred to~\cite{hutter_automated_2019}
for more information.

There are several other approaches that provide a high-level language for the
configuration of a software system that allows expression of common solution
approaches for a particular problem, in particular in AI.
Multi-TAC~\cite{minton_automatically_1996} provides high-level LISP constructs
to solve constraint satisfaction problems (CSP), allowing expression of different
heuristics and types of CSPs to be solved. The KIDS
system~\cite{westfold_synthesis_2001} provides similar functionality and in
addition reformulates the problem to solve it more efficiently. The Dominion
architecture \cite{balasubramaniam_dominion_2011} allows generating C++ CSP
solvers that are specialized to a particular type of problem based on code
templates. Aeon~\cite{monette_aeon_2009} allows for the flexible construction of
solvers for scheduling problems. The Heuristic Search
Framework~\cite{dorne_hsf_2001} provides building blocks for heuristic search
that can be assembled based on the problem to be solved.
PetaBricks~\cite{ansel_petabricks_2009} is not specific to any particular type
of problem and allows expressing and configuring algorithmic choices more
generally.

SATenstein~\cite{khudabukhsh_satenstein_2009} provides components for solvers
for the satisfiability problem that do not have to be assembled explicitly, but
are selected by setting hyperparameters of the software. The authors use the
SMAC~\cite{hutter11smac} hyperparameter tuner to automatically tailor SATenstein
to particular sets of problems. Similarly,~\cite{LopStu2012tec} use
irace~\cite{lopez2016irace} to automatically configure an ant colony
optimization algorithm with multiple competing objectives. We take a similar
approach here in that the algorithmic choices are exposed as hyperparameters
that can be tuned without the need for an approach that is specialized to our
setting.

One of the problems with automatically designing software from building blocks
is that not all combinations of building blocks are necessarily valid or solve
the target problem. Common solutions to this problem are to make all combinations
valid, or to automatically identify and prohibit/penalize invalid combinations. A
particularly elegant solution is presented in~\cite{seipp_new_2015} in the
context of search heuristics, where declarative constraints specify what
components can be combined and how they can be combined. Every solution to the combinatorial problem defined by these constraints represents a valid design. In this paper, we take a
similar approach -- all configurations are a valid design.

The general approach of allowing algorithmic choices in a software system
instead of fixing them at implementation time has been termed ``Programming by
Optimization''~\cite{hoos_programming_2012} (PBO).
HPO can be seen as an instance of PBO in a certain sense, but 
we are not aware of many cases where PBO is applied to designing HPO systems themselves. 
There are many approaches that
focus on individual algorithmic choices (e.g.\ the choice of surrogate model for
Bayesian optimization~\cite{malkomes2018automating}) without the flexibility of
our approach, or optimizing a meta-algorithmic process, e.g.\ \cite{lindauer2019assessing} investigate the impact of tuning the hyperparameters of a Bayesian optimization process and \cite{metairace} configure irace on a set of surrogate benchmarks.

Most existing approaches do not analyze the algorithmic choices of an optimized
system, and the ones that do perform relatively straightforward analyses. For
example, \cite{minton_automatically_1996}~compare the designs their approach
finds automatically to the designs expert humans generated.
\cite{fawcett2016analysisAblation}~evaluate the contribution of one component at
a time to the change between a default and an optimized design in order to automatically
identify the most important change. \cite{moaco}~perform ANOVA and
non-parametric Friedman tests to investigate in detail the effects that algorithmic
choices have for ant colony optimization algorithms.


%% file: 25_methodology.tex
\subsection{Supervised Machine Learning}
Supervised ML typically deals with a dataset (which is, mathematically speaking, a tuple) $\D = \left((\xi, \yi)\right) \in (\Xspace \times \Yspace)^n$ of $n$ observations, 
assumed to be drawn i.i.d.\ from  a data-generating distribution $\Pxy$. 
An ML model is a function $\fh:\Xspace \rightarrow \R^g$ that assigns a prediction to a feature vector from $\Xspace$. 
$\fh$ is itself constructed by an \emph{inducer} function $\inducer$, i.e., the model-fitting algorithm. The inducer $\inducer : (\D, \lambdav) \mapsto \fh$  uses training data $\D$ and a vector of \emph{hyperparameters} $\lambdav$ that govern its behavior. 
The overall goal of supervised ML is to derive a model $\fh$ from a data set $\D$, so that $\fh$ predicts data sampled from $\Pxy$ best. 
The quality of a prediction is measured as the discrepancy between predictions and ground truth. This is operationalized by the loss function $L : \Yspace \times \R^g \rightarrow \R^+_0$, which is to be minimized during model fitting. 
The expectation of the loss value of predictions made for (unseen) data samples drawn from $\Pxy$ is the \emph{generalization error}
\begin{equation}\label{eqn:ge}
    GE := \E_{(\xv, y) \sim \Pxy} \left[L(y, \fh(x))\right] 
\end{equation}
which can not be computed directly if $\Pxy$ is not known beyond the available data $\D$. 
One therefore often uses so-called \emph{resampling} techniques~\cite{bischl_evco12a} that fit models on $N_\mathrm{iter}$ subsamples $\D[J_j]$ and evaluate them on complements $\D[-J_j]$ of these subsets to obtain an estimate of the generalization error
\begin{equation}\label{eqn:resample}
\widehat{GE}(\inducer, \lambdav, \mathbf{J}) =
  \frac{1}{N_\mathrm{iter}}\sum_{j=1}^{N_\mathrm{iter}} L\big(y[-J_j], \inducer\left(\D[J_j], \lambdav\right)(x[-J_j])\big)\textrm{.}
\end{equation}

Depending on the resampling method, the inducer $\inducer$, and the quantity of data in $\D$, evaluating the resampling objective in Equation~\ref{eqn:resample} can require large amounts of computational resources.

\subsection{Hyperparameter Optimization (HPO)}

The goal of HPO is to identify a hyperparameter configuration that performs well in terms of the (estimated) generalization error in Eq.~\ref{eqn:resample}.
Often, optimization only concerns a subspace of available hyperparameters because some hyperparameters might be set based on prior knowledge or due to other constraints. Formally, one can split up the space of hyperparameters $\Lambda$ into a subspace of hyperparameters influencing the performance $\Lambda_S$ and the remaining hyperparameters $\Lambda_C$. We define the HPO problem as:
\begin{eqnarray}
    \lambdav_S^\ast \in \argmin_{\lambdav_S \in \Lambda_S} c(\lambdav_S) = \argmin_{\lambdav_S \in \Lambda_S} \widehat{GE}(\inducer, (\lambdav_S, \lambdav_C), \mathbf{J}). 
    \label{eq:hpo_objective}
\end{eqnarray}
Here, $\lambdav^\ast_S$ denotes the theoretical optimum and $c(\lambdav_S)$ is a shorthand for the estimated generalization error in Eq.~\eqref{eqn:resample}. 

Hyperparameters can be either continuous, discrete, or categorical, and search spaces are often a mix of the different types. The search space may be hierarchical, i.e., some (child) hyperparameters can only be set in a meaningful way if another (parent) hyperparameter takes a certain value. Therefore, the search space $\Lambda_S$ might be mixed and hierarchical. 
In particular, many AutoML frameworks perform optimization over a hierarchical hyperparameter space that represents the components of a complex ML pipeline~\cite{bischl21hpo,hutter_automated_2019,kotthoff2019autoweka}. 


Many HPO algorithms can be characterized by how they handle two different trade-offs: 
a) The exploration vs. exploitation trade-off refers to how much budget an optimizer spends on either trying to directly exploit the currently available knowledge base by evaluating very close to the currently best candidates (e.g., local search) or whether it explores the search space to gather new knowledge (e.g., random search).
b) The inference vs. search trade-off refers to how much time and overhead is spent to induce a model from the currently available archive data in order to exploit past evaluations as much as possible.
Other relevant aspects that HPO algorithms differ in are: \textit{Parallelizability}, i.e. how many configurations a tuner can (reasonably) propose at the same time; \textit{global vs. local} behavior of the optimizer, i.e. if updates are always quite close to already evaluated configurations; \textit{noise handling}, i.e., if the optimizer takes into account that the estimated generalization error is noisy; \textit{multifidelity}, i.e., if the tuner uses cheaper evaluations, for example on smaller subsets of the data, to infer performance on the full data; \textit{search space complexity}, i.e., if and how hierarchical search spaces can be handled.

\subsection{Multifidelity HPO}

The HPO problem defined in~\eqref{eq:hpo_objective} is challenging. One of the reasons is that the value of the objective function $c(\lambdav_S)$ is expensive to evaluate. Optimization methods for HPO attempt to overcome this in two ways: by being very sample-efficient and attempting to extract as much information from previous objective evaluations as possible, and by using \emph{multifidelity methods} that evaluate cheaper proxy functions of $\widehat{GE}$.

Multifidelity methods make use of the fact that the resampling procedure in Equation~\eqref{eqn:resample} can be modified in multiple ways to make evaluation cheaper:
one can
(i)~reduce the training sizes $\left|J_j\right|$ via subsampling, as model evaluation complexity is often at least linear in training set size,
or (ii)~change some components in $\lambdav$ in a way that makes model fits cheaper, such as reducing the overall number of training cycles performed by a neural network fitting process, or reducing the number of base learner fits in a bagging or boosting method.
These modifications can both increase the variance of $\widehat{GE}$ and also introduce an (often pessimistic) bias, as models trained on smaller datasets or with values of $\lambdav$ that make fitting cheaper often have worse generalization error.

We introduce a \emph{fidelity} hyperparameter $r\in (0,1]$ that influences the resource requirements of the evaluation of $\widehat{GE}$  and thus the overall budget spent for the optimization. It can influence the size of the training dataset or the $\lambdav$ to trade off between evaluation efficiency and information gained about the generalization error. In the latter case, $r$ can only influence hyperparameters separate from $\lambdav_S$, i.e., $\lambdav_C$. Typically, $r$ only affects one of these aspects at a time, and if it affects $\lambdav_C$, it only affects a single hyperparameter dimension. We define
\begin{equation}
c(\lambdav_S; r) := \widehat{GE}\left(\inducer, \left(\lambdav_S, \lambdav_C(r)\right), \mathbf{J}\right). 
\end{equation}
If the assumption holds that a higher fidelity $r$ returns a better model in terms of the generalization error (e.g., if $r$ controls the size of the training set via sub-sampling), we are interested in finding the configuration $\lambdav_S$ for which the inducer will return the best model given the full budget $r=1$. Under this assumption, we define $c(\lambdav_S) := c(\lambdav_S; 1)$, see e.g.\ \cite{klein2020model}, and define the optimization problem as in Equation \eqref{eq:hpo_objective}. In practice, an HPO algorithm would return a configuration $\lambdav_S^\ast$. A final model might be trained for $\lambdav_S^\ast$ for the full fidelity $r = 1$ after optimization (if not yet done by the optimizer). This assumption may be violated in some scenarios and model performance could worsen for a higher fidelity (e.g., a neural network, which may overfit on a small dataset if trained for too many epochs). In this case, we define the optimization problem as $(\lambdav_S^\ast, r^\ast) \in \text{arg min}_{\lambda_S \in \Lambda_S, r \in (0, 1]} c(\lambda_S; r)$. 

The resource requirements of evaluating $c(\lambdav; r)$ can have a complicated relationship with $\lambdav$ and $r$; in practice, $r$ is chosen such that it has an overwhelming and linear influence on resource demand.
The overall cost of optimization up to a point in time is therefore assumed to be the cumulative sum of the values of $r$ of all evaluations of $c(\lambdav; r)$ up to that point.
We can also interpret $r$ as the fraction of the overall budget that has been spent in this way.

Given the definition of the HPO problem, we present a (multifidelity) hyperparameter optimization algorithm for a single worker in its most generic form (see Alg.~\ref{alg:multifid}).  
Until a pre-determined budget is exhausted, such an algorithm decides in every iteration (a) which configuration(s) $\lambdav_S$ to evaluate next and (b) which fraction of the budget $r$ to allocate (non-multifidelity algorithms set this to $r = 1$ as default). The algorithm makes use of an \emph{archive} $\archive$, a database recording previously proposed hyperparameter configurations and, if available, their evaluation results.

\begin{algorithm}[H]
\caption{A generic HPO algorithm}
\label{alg:multifid}
\begin{algorithmic}[1]
    \While{budget is not exhausted}
        \State Propose $\left(\lambdav_S^{(i)}, r^{(i)}\right), i = 1, ..., k$, based on archive $\archive$ \label{algline:propose}
        \State Write proposals into a shared archive $\archive$
        \State Evaluate configuration(s) $c\left(\lambdav_S^{(i)}; r^{(i)}\right)$
        \label{algline:evaluate}
        \State Write results into shared archive $\archive$
    \EndWhile
    \State Wait for workers to synchronize 
    \State Return best configuration in archive $\archive$
\end{algorithmic}
\end{algorithm}

The optimization process can be accelerated by making efficient use of parallel resources. We distinguish between \emph{synchronous} and \emph{asynchronous} scheduling. The former starts multiple evaluations synchronously at the same time. To be more precise, a number of $k > 1$ configurations is proposed in line~\ref{algline:propose} and evaluated in parallel in line~\ref{algline:evaluate}, all within the inner loop of Alg.~\ref{alg:multifid}. Given $K$ available parallel resources, it should be ensured that the number $k$ of configurations that are scheduled in parallel is not significantly smaller than $K$, and that the evaluation runtimes amongst these $k$ configurations do not differ significantly in order to avoid idling single parallel resources. In contrast, for asynchronous scheduling, Alg. \ref{alg:multifid} is individually run on a set of $K$ different worker resources. Given a shared archive that is synchronized amongst the workers, every worker can independently schedule evaluations as soon as they are free. 

\empty

Common established HPO algorithms are instances of Alg.~\ref{alg:multifid} as described below and summarized in Table~\ref{tab:algorithms}.

\subsubsection{Random Search}
Configurations are drawn (uniformly) at random, and every configuration is evaluated with full fidelity $r = 1$. Parallelization is straightforward, as configurations are drawn independently. 

\subsubsection{Bayesian optimization}
\label{sec:smbo}
In Bayesian optimization (BO), the configuration that maximizes an acquisition function $a(\lambdav)$ (e.g., expected improvement, EI~\cite{jones98ego}, or the lower confidence bound, LCB) is proposed and evaluated with the full fidelity $r = 1$. $a(\lambdav)$ is based on a surrogate model trained on the archive $\archive$. BO can be parallelized by either using methods that can propose multiple points at the same time using a single surrogate model, or alternatively by fitting a surrogate model on the anticipated outcome of configurations that were proposed but not yet evaluated~\cite{bischl14moimbo}.

\subsubsection{FABOLAS}
Fabolas~\cite{klein2017fast} is a continuous multi-fidelity BO method based on BO, where the conditional validation error is modelled as a Gaussian process using a complex kernel capturing covariance with the training set fraction $r \in (0,1]$ to allow for adaptive evaluation at different resource levels.

\subsubsection{Successive Halving}

Successive halving (SH)~\cite{jamieson16successivehalving} is a simple multi-fidelity optimization algorithm that combines random sampling of configurations with a fixed schedule for $r$. At the beginning, a batch of $\mu$ configurations is sampled randomly and evaluated with an initial fidelity $r_\text{min} < 1$. This is followed by repeated ``halving'' steps, where the top fraction $\eta^{-1}$ of configurations is kept and evaluated after $r$ is increased by a factor of $\eta$, until the maximum fidelity value is reached. The schedule is chosen to keep the total sum of all evaluated $r$ constant in each batch.

\subsubsection{Hyperband}
Similar to SH, Hyperband~\cite{li2017hyperband} uses a fixed schedule for the fidelity parameter $r$, but it augments SH by using multiple \emph{brackets} $b$ of SH runs starting at different $r_\text{min}(b)$ and with different $\mu(b)$.

The number of brackets is set to $s = \lfloor - \log_\eta (r_\text{min})\rfloor + 1$, which coincides with the number of fidelity steps that can be performed on a geometric scale on the interval $[r_\text{min}, 1]$. In bracket $b \in \{1, 2, \ldots , s\}$, a number of $\mu(b) = \lceil s \cdot \frac{\eta^{s-b}}{s-b+1}\rceil$ are initially sampled and evaluated with initial fidelity $r = \eta^{s-b}$. $\mu(b)$ is chosen such that each bracket approximately spends a similar overall budget. 

\subsubsection{Asynchronous Successive Halving (ASHA) and Asynchronous Hyperband}

Hyperband, as well as SH, have the drawback that batch sizes decrease throughout the stages of an SH run, preventing efficient utilization of parallel resources.
ASHA \cite{li20asha} is an effective method to parallelize SH by an asynchronous parallelization scheme. A shared archive across a number of different workers is maintained. Instead of waiting until all $n$ configurations of a batch have been evaluated for fidelity $r$, every free worker queries the shared archive $\archive$ for promotable configurations (i.e., configurations that belong to the fraction of top $\eta^{-1}$ configurations evaluated with the same fidelity). Asynchronous Hyperband works similarly.

\subsubsection{Bayesian Optimization Hyperband (BOHB)}
\label{sssec:bohb}

Model-based methods outperform Hyperband when a relatively large amount of budget is available and many objective function evaluations can be performed. BOHB~\cite{falkner18bohb} was created to overcome this drawback. It iterates through successive halving brackets like Hyperband, but, instead of sampling new configurations randomly, it uses information from the archive to propose points that are likely to perform well. $N_s$ total configurations are proposed for evaluation; $\rho$ are sampled at random and the rest are chosen based on Bayesian optimization with a surrogate model induced on the evaluated configurations in $\archive$. The models used by BOHB are a pair of kernel density estimators of the top and bottom configurations in $\archive$, similar to the process in~\cite{bergstra11treeparzen}. 

\subsubsection{A-BOHB}
\cite{klein20abohb} propose A-HBOB, an asynchronous extension of BOHB where configurations are sampled from a joint Gaussian Process, explicitly capturing correlations across fidelities. In contrast to ASHA~\cite{li20asha} and asynchronous versions of BOHB implemented by~\cite{falkner18bohb}, A-HBOB does not perform synchronization after each stage but instead uses a stopping rule~\cite{golovin2017google} to asynchronously determine whether a configuration should be promoted to the next rung (and immediately scheduled) or terminated.

%% file: 30_multifid_hpo.tex
We aim to find an HPO algorithm that performs particularly well in the multifidelity setting. To design an algorithm by optimization, we propose a framework and search space of HPO algorithm candidates that covers a large class of possible algorithms as in Alg.~\ref{alg:multifid}, and focus on a subclass of algorithms similar to Hyperband because of their favorable properties. This subclass is limited to multifidelity algorithms that use a pre-defined schedule of geometrically increasing fidelity evaluations, and therefore contains fixed-fidelity-schedule HPO algorithms like BOHB and Hyperband. 

The basis of this framework is presented as Algorithm~\ref{alg:smashy}, which has configuration parameters that allow it to be tailored to specific benchmarks by combining algorithmic building blocks in novel ways. The main difference to Alg.~\ref{alg:multifid} is that the \textit{Propose} part is specified more explicitly. At its core, Alg.~\ref{alg:smashy} consists of two parts: (i)~sampling new configurations at low fidelities (lines~\ref{algline:startsample}--\ref{algline:endsample}) and (ii)~increasing the fidelity for existing configurations (lines~\ref{algline:startprogress}--\ref{algline:endprogress}).  
In contrast to Alg.~\ref{alg:multifid}, Alg.~\ref{alg:smashy} makes use of state variables $t$, $b$, and $r$ to account for optimization progress. However, these variables are only shown in Alg.~\ref{alg:smashy} for clarity and can, in principle, be inferred from the archive~$\archive$. As argued in Section~\ref{sec:25_methodology}, every single worker instance of Alg.~\ref{alg:multifid} can in principle be scheduled asynchronously, but we do not consider this in this work. 

First, Alg.~\ref{alg:smashy} uses a $\textsc{Sample}$-subroutine to initialize the initial batch $C$ of $\mu$ solution candidates. The fidelity of the evaluation of the proposed configurations is refined iteratively; when all configurations in the batch have been evaluated with given fidelity $r$, the top $1 / \eta_\text{surv}$ fraction of configurations is evaluated with a fidelity that is increased by a factor of $\eta_\text{fid}$. When the fidelity cannot be further increased for a batch because all of its configurations were evaluated at full fidelity $r = 1$, they are set aside, and a new batch of configurations is sampled.


The \textsc{Sample} subroutine creates new configurations to be evaluated, possibly using information from the archive to propose points that are likely to perform well, similar to BOHB. Instead of using a pair of kernel density estimators, we allow that any inducer $\inducer_{f_\text{sur}}$ that produces a surrogate model $f_\text{sur}$ can be used for model-assisted sampling\footnote{BOHB itself uses an inducer that produces a function to calculate the ratio of kernel densities, an unusual kind of regression model.}. The subroutine works by at first sampling a number of points from a given generating distribution $\P_{\lambdav}(\archive)$. The performance of these points is then predicted using the surrogate model, and points with unfavorable predictions are discarded a process which we refer to as \textit{filtering}. This process is repeated until the requested number $\mu$ of points that were not discarded is obtained. $N_\text{s}$ and $\rho$ have the same function as in~\cite{falkner18bohb} (see Section~\ref{sssec:bohb}), with $N_\text{s}$ controlling the number of sampled points needed for each of the $\mu$ points returned, and $\rho$ controlling the fraction of points that are not filtered. Thus the configuration space of sampling methods also includes purely random sampling, as in Hyperband, by setting $\rho=1$. The influence of the surrogate model on sampled candidates is larger when (i) the number of sampled configurations $N_\text{s}$ is large, or (ii) the fraction of candidates sampled at random $\rho$ is small. We present two slightly different \textsc{Sample} algorithms (Algorithms \ref{alg:tournament} and~\ref{alg:progressive}) based on this principle.

Algorithm \textsc{SampleTournament} (pseudocode in Appendix~\ref{app:algs}, Alg.~\ref{alg:tournament}) diversifies the set of points proposed through an extension that uses different values of $N_\text{s}^{(i)}$ for different points $\lambdav_S^{(i)}, i \in \{1, \ldots, \mu\}$ sampled at each invocation of \textsc{Sample}. We parameterize this using the configuration parameters $N_\text{s}^0$ and $N_\text{s}^1$; the effective value of $N_\text{s}$ for each point is interpolated geometrically. This means that, when $n$ points are sampled, the $i$th point is chosen from a set of $\left\lfloor(N_\text{s}^0)^{(n-i)/\mu}\cdot (N_\text{s}^1)^{i/\mu}\right\rceil$ randomly sampled points. We furthermore make it possible to draw more than one point from the same independent random search sample: instead of drawing one point from $N_\text{s}^{(i)}$ samples, we draw $n_\textrm{per\_tournament}$ points from $n_\textrm{per\_tournament}\cdot N_\text{s}^{(i)}$ samples. 

Besides the sampling method described above, we propose an alternative method, which we name \textsc{SampleProgressive} (pseudocode in Appendix~\ref{app:algs}, Alg.~\ref{alg:progressive}): instead of sampling $N_\text{s}(i)$ points independently for each configuration $\lambdav_S^{(i)}$, we sample a single ordered pool $\mathcal{P}$ of $\mu\cdot\textrm{max}(N_\text{s}^0, N_\text{s}^1)$ random points once at the beginning of \textsc{Sample}. Each $\lambdav_S^{(i)}$ is then selected as the point with the best surrogate-predicted performance from the first $\mu\cdot N_\text{s}(i)$ points in $\mathcal{P}$ that was not already selected before. 

While hyperparameters $\lambdav_S$ are proposed by one of the two \textsc{Sample} methods, the fidelity hyperparameter $r$ follows a fixed schedule similar to SH and Hyperband, with a few extensions. For one, the fraction of survivors $\eta_\text{surv}$ can be a different value from the fidelity scaling factor $\eta_{fid}$. Furthermore, the algorithm allows two scheduling modes, controlled by \emph{batch\_method}: The \texttt{HB} mode evaluates brackets, as performed by Hyperband. $\mu(b)$ is then adjusted to make total budget expenditure approximately equal between all brackets, and depends on both $\eta_\text{surv}$ and $\eta_{fid}$ in a more complex way than in Hyperband. On the other hand, the \texttt{equal} \emph{batch\_method} uses equal batch sizes for every evaluation. Individuals that perform badly at low fidelity are removed, as in SH, but new individuals are sampled to fill up batches to the original size. Because new individuals are added to the batches at all fidelity steps, it is not necessary to use brackets with different initial fidelities, and therefore only a single repeating bracket $b=1$ is used.


If the exploration-exploitation tradeoff is not balanced properly, optimization progress can either stagnate or functions evaluation are wasted due to too much exploration of uninteresting regions of the search space.
However, the relative importance of exploration and exploitation can change throughout the course of optimization, where exploration performed later during the optimization is not as useful as during the beginning.
The given configuration space makes it possible to make the exploration-exploitation tradeoff dependent on optimization progress by providing the option to make $\rho(t)$ and $\left(N_\text{s}^0(t), N_\text{s}^1(t)\right)$ dependent on the proportion of exhausted total budget at every configuration proposal step. It is likely that large values of $\rho(t)$ / small values of $N_\text{s}(t)$ perform better when $t$ is small, and conversely, small $\rho(t)$ / large $N_\text{s}(t)$ work well for large $t$.

\begin{algorithm}[t]
\caption{\textsc{Smashy} algorithm}
\label{alg:smashy}
    \hspace*{\algorithmicindent} \textbf{Configuration Parameters}: batch size schedule $\mu(b)$, number of fidelity stages $s$, 
    survival rate $\eta_\text{surv}$, fidelity rate $\eta_\text{fid}$, \textsc{Sample} method (either \textsc{SampleTournament} or \textsc{SampleProgressive}), $\textit{batch\_method}$ (either \texttt{equal} or \texttt{HB}), total budget $B$; further configuration parameters of \textsc{Sample}: $\inducer_{f_\text{surr}}$, $\P_{\lambdav}(\archive)$, $\rho(t)$, $\left(N_\text{s}^0, N_\text{s}^1(t)\right)$, $n_\mathrm{trn}$.
    
    \vspace*{0.2cm}
    \hspace*{\algorithmicindent} \textbf{State Variables}: Expended budget fraction $t \leftarrow 0$, bracket counter $b \leftarrow 1$ (remains 1 for \textit{batch\_method} = \texttt{equal}), current fidelity $r \leftarrow 1$, batch of proposed configurations $C \leftarrow \emptyset$
\begin{algorithmic}[1]
    \vspace*{0.2cm}
    \While{$t < 1$}\vspace*{0.2cm}

        \If{$r = 1$}\label{algline:startsample}
            \Comment{Generate new batch of configurations}
            \State $r \leftarrow (\eta_\text{fid})^{b - s}$
            \State $C \leftarrow \textsc{Sample}\big($
                    \parbox[t]{2cm}{
                    $\archive, \mu(b), r; \inducer_{f_\text{sur}}, \P_{\lambdav}(\archive),$ \\
                    $\rho(t), \left(N_\text{s}^0(t), N_\text{s}^1(t)\right), n_\mathrm{trn}\big)$
                    }
                    
            \If{\textit{batch\_method =} \texttt{HB}}
                \State $b \leftarrow (b\,\mathrm{mod}\,s) + 1$
            \EndIf\label{algline:endsample}
        \Else
            \Comment{Progress fidelity}\label{algline:startprogress}
            \State $r \leftarrow r \cdot \eta_\text{fid}$
            \State $C \leftarrow \textsc{select\_top}\left(C, |C| / \eta_\text{surv}\right)$
            \If{\textit{batch\_method = } \texttt{equal}}
                \State $\tilde \mu \leftarrow \mu(b) - |C|$
                \State $C \leftarrow C \cup \textsc{Sample}\big($\parbox[t]{2cm}{$
                    \archive, \tilde \mu, r; \inducer_{f_\text{sur}}, \P_{\lambdav}(\archive), \\
                     \rho(t), \left(N_\text{s}^0(t), N_\text{s}^1(t)\right), n_\mathrm{trn}\big)$}
                    
            \EndIf\label{algline:endprogress}
        \EndIf\vspace*{0.2cm}
        \State Evaluate configuration(s) $c\left(\lambdav_S; r\right)$ for all $\lambdav_S \in C$
        \State Write results into shared archive $\archive$
        \State $t \leftarrow t + r \cdot |C| / B$
        \Comment{Update budget spent}
    \EndWhile
    \vspace*{0.2cm}
\end{algorithmic}
    \hspace*{\algorithmicindent}* possibly adapted during the course of optimization

\end{algorithm}

Common HPO algorithms (\textsc{RS}, \textsc{BO}, \textsc{SH}, \textsc{HB}, \textsc{BOHB}) can be instantiated within this framework (see Table \ref{tab:algorithms}). 

\begin{table*}[htbp]
\caption{\textsc{RS}, \textsc{BO}, \textsc{HB}, \textsc{BOHB} as instances of Algorithm \ref{alg:smashy}. $\eta$, $\rho$, $N_\text{s}$ are configuration parameters of the respective algorithms. \\
``---'' denotes that the value has no influence on the algorithm in this configuration.\\ *: BO and BOHB use inducers that produce non-standard model functions, which do not aim to predict the actual outcome of configurations, and instead calculate the value of an acquisition function such as EI~\cite{jones98ego} (for BO) or the ratio of two KDE models (for BOHB).\\ $\dagger{}$: In a small departure from BOHB, Alg.~\ref{alg:smashy} uses the KDE estimate of good points for all sampled points, even when randomly interleaved. BOHB does random interleaving from a uniform distribution.}
\centering
\begin{tabular}{lccccccccc}
\toprule
Algorithm & $\mu(b)$ & $s$ & $\eta_\text{surv}$ & $\eta_\text{budget}$ & $\inducer_{f_\text{sur}}$ & $\rho$ & $N_\text{s}$ & \emph{batch\_mode} & $\P_{\lambdav}(\archive)$ \\ \midrule
RS & --- & 1 & --- & --- & --- & 1 & --- & --- & uniform \\
BO & 1   & 1 & --- & --- & e.g.\ GP+EI* & $\rho$ & $N_\text{s}$ & --- & uniform \\
HB & $\lceil s \cdot \frac{\eta^{s-b}}{s-b+1}\rceil$   & $\lfloor - \log_\eta (r_\text{min})\rfloor + 1$ & $\eta$ & $\eta$ & --- & 1 & --- & \textsc{HB} & uniform \\
BOHB & $\lceil s \cdot \frac{\eta^{s-b}}{s-b+1}\rceil$ & $\lfloor - \log_\eta (r_\text{min})\rfloor + 1$ & $\eta$ & $\eta$ & TPE* & $\rho$ & $N_\text{s}$ & \textsc{HB} & KDE${}^\dagger$ \\
\bottomrule
\end{tabular}
\label{tab:algorithms}
\end{table*}




%% file: 35_design_and_analysis.tex
\subsection{Algorithm Design}
\newcommand{\gammav}{\bm{\gamma}}

Our goal is to configure a new HPO algorithm based on a superset of design choices included in previously published HPO methods. We are interested in finding a configuration (or making design choices) based on a set of training instances that works across a broad set of future problem instances. This problem is called \textit{algorithm configuration} \cite{birattari2009tuning, hutter11smac}. It is quite similar to HPO, while a major usual difference is that algorithm configuration optimizes the configuration of an arbitrary algorithm over a diverse set of often heterogeneous instances for optimal average performance, HPO performs a per-instance configuration of an ML inducer for a single data set.
We introduce the following notation for consistency with the relevant literature: $\gammav$ denotes configuration-parameters controlling our optimizer $A$, while $\lambdav$ denotes hyperparameters optimized by our optimizer (controlling our inducer $\inducer$).
The problem can be formally stated as follows: Given an algorithm $A: \Omega \times \Gamma \rightarrow \Lambda$ parametrized by $\gammav \in \Gamma$ and a distribution $\P_\Omega$ over problem
instances $\Omega$ together with a cost metric $\zeta$, we must find a parameter setting $\gammav^\ast$ that minimizes the expected $\zeta(A)$ over $\P_\Omega$:
\begin{eqnarray}
    \gammav^\ast \in \argmin_{\gammav \in \Gamma} \E_{\omega \sim \P_\Omega} \left[ \zeta(A(\omega, \gammav)) \right]\mathrm{.}
    \label{eq:ac_objective}
\end{eqnarray}
In our example, $\Gamma$ corresponds to the space of possible components of our HPO method and $\Omega$ to a class of HPO problems (i.e.\ ML methods and datasets on which they are evaluated) for which their configuration should be optimal.
Based on a training set of representative instances $\{\omega_i\}$ drawn from $\P_\Omega$, a configuration $\gammav^\ast$ that minimizes $c$ across these instances should be chosen through optimization. Here we refer to this process as \emph{meta-optimization} if necessary to distinguish it from the optimization performed by the HPO algorithm $A$.

A variety of racing-based strategies have been proposed for algorithm configuration problems, such as F-RACE~\cite{birattari2010f} and IRACE~\cite{lopez2016irace}, along with non-parametric alternatives like ParamILS~\cite{hutter2009paramils} as well as genetic algorithm-based variants (e.g.\ GGA~\cite{ansotegui2009gender}) and sequential model-based optimization (SMBO) variants based on Bayesian optimization (such as SMAC~\cite{hutter11smac}). We describe two common algorithm configuration methods (SMAC, IRACE) below.

SMAC~\cite{hutter_automated_2019} extends the SMBO paradigm described in Section~\ref{sec:smbo} to an algorithm configuration setting. This is achieved through the use of an intensification procedure that governs across how many problem instances each configuration is evaluated, trading off computational cost against confidence regarding the superiority of a given configuration. Furthermore, instance features $\mathrm{if}(\omega)$ describing properties of $\omega$ are used to train the empirical performance model (EPM) $\mathrm{if}(\omega) \times \gammav \mapsto \hat\zeta$ predicting the performance $\zeta(A(\omega, \gammav)$ of a configuration $\gammav$ on a new problem instance $\omega$. IRACE~\cite{lopez2016irace} does not make use of an EPM, but instead iteratively repeats the following steps: After sampling new candidate configurations according to a sampling distribution, a set of elite configurations are selected through a racing procedure~\cite{maron1997racing}. These configurations are used to update the sampling distribution in order to focus exploration around the current set of elite configurations. During racing, statistical tests are used to preemptively terminate unpromising configurations with high cost $\zeta$ across instances.

\subsection{Algorithm Analysis}

The design of well-performing algorithms is often the focus of research. However, gaining an understanding of the effect of the design choices that achieve this performance is crucial for multiple reasons. 
First, this understanding of such effects is directly interesting in itself.
Second, design by optimization carries the risk of producing overly complex algorithms; if algorithmic elements have only minor or no effect on algorithm performance they may not be necessary. 
Third, sophisticated algorithmic components might be praised as drivers of algorithmic performance, while the actual importance of minor modifications is not properly highlighted. 
Hence, it is crucial for research to detect and analyse the effects of all components to be able to direct future efforts at improvement. 

The field of \emph{sensitivity analysis} (SA) -- which generally deals with assessing which inputs (to a mathematical model) determine the variance of an output -- comprises a multitude of methods to assess importance of input factors on the output of a mathematical model~\cite{saltelli02saimportance}. 
Commonly used methods are analysis of variance (ANOVA) and functional ANOVA (fANOVA) methods, which decompose the response of a (mathematical) model or function into lower-order components~\cite{fawcett2016analysisAblation}. In the field of ML, fANOVA is a popular way of analyzing the importance of hyperparameters~\cite{hutter14fanova}. 

Performing a sensitivity analysis can be computationally expensive, as it usually requires evaluation of the mathematical model (in our case, running an HPO algorithm) on some experimental design. We want to limit the costs of performing such a sensitivity analysis, either by approximating the mathematical model to be investigated by a surrogate~\cite{cheng2020surrogate}, or by choosing a method that keeps the resource requirements of the experimental design reasonably low, like for example a \emph{one-factor-at-a-time} (OFAT)~\cite{sheikholeslami21autoablation} analysis. Those methods have their own limitations; for example, any interactions between inputs cannot be detected by an OFAT analysis, and parts of the input space remain hidden. 

A popular way of performing SA in practice are \emph{ablation} studies. An ablation study involves measuring the performance of a mathematical model (e.g.\ the performance of a deep neural network) when removing one or more of its components (e.g.\ layers) to understand the relative contribution of the ablated components to overall performance~\cite{sheikholeslami21autoablation}. There are different ways of performing an ablation analysis; probably the most common approach is \emph{Leave-One-Component-Out} (LOCO) ablation~\cite{sheikholeslami21autoablation}. 

In the context of algorithm configuration, \cite{fawcett2016analysisAblation} propose an ablation approach that links a source configuration (e.g.\ the default) to a target (e.g.\ the optimized configuration) through an ablation path.

%% file: 40_experimental_design.tex
\subsection{Benchmark Setup}

Given the formalization of the generic MF-HPO algorithm in Section~\ref{sec:30_multifid_hpo}, our goal is to (1)~find the best representative (out of this class of algorithms) via optimization, and (2)~explain the role of specific algorithmic components in a benchmark-driven approach. To allow for meaningful conclusions on both aspects, a benchmark problem must be carefully selected to represent a relevant set of optimization instances on which each candidate algorithm is run and evaluated.



We rely on benchmark scenarios of the YAHPO Gym benchmark suite~\cite{pfisterer_yahpo_2021}, each of which provides a number of related \emph{instances} of optimization problems. Benchmarks in the YAHPO Gym are implemented as surrogate model benchmarks, where a Wide \& Deep~\cite{cheng2016wide} neural network was fitted to a set of pre-evaluated performance values of hyperparameter configurations. These have the advantage that they can be evaluated quickly, making them well-suited for meta-optimization, with a more realistic objective landscape than synthetic functions or benchmarks based on lookup tables.

Benchmarks are useful both for (meta-)optimization and for comparison of different methods. However, when comparing the performance of an algorithm found by meta-optimization with other methods, it is necessary to make this comparison on instances that differ from the instances used for meta-optimization. This is to avoid an optimistic bias caused by ``overfitting'' to random peculiarities of the particular instances used for optimization that are not representative of the class of optimization problems as a whole. YAHPO Gym provides a set of dedicated ``training'' instances for meta-optimization, and a set of ``test'' instances for evaluation of algorithmic performance. 

The benchmark collections we have chosen cover three important application areas of AutoML: Hyperparameter optimization, AutoML pipeline configuration, and neural architecture search (NAS). These classes of problems do not only represent common and relevant tasks for researchers and practitioners in the field; as presented in Table~\ref{table:benchmark_instances}, they are also quite different regarding: (1)~the dimensionality of the search space, (2)~the types of the hyperparameters (categorical, integer, continuous) in the search space, and (3)~whether there are hierarchical dependencies between hyperparameters in the search space.  While the underlying data for lcbench and nb301 have been previously used in publications \cite{ZimLin2021a,siems2020nasbench301}, rbv2\_super is a novel task that has not been investigated previously in literature.

\vspace{\baselineskip}
\textbf{HPO on a neural network:} The first set of problems, \emph{lcbench} \cite{ZimLin2021a}, covers hyperparameter optimization (HPO) on a relatively small and numeric search space. The neural network (more precisely, a funnel-shaped multilayer perceptron) that is tuned has a total of $7$ numerical hyperparameters. The fidelity of an evaluation can be controlled by setting the number of epochs over which the neural network is trained. The instances belonging to this scenario represent HPO performed on $35$ different classification tasks taken from OpenML~\cite{vanschoren2013openML}. As a target metric, we choose the cross entropy loss on the validation set.

\textbf{AutoML pipeline configuration:} Second, we investigate the problem of configuring an AutoML pipeline. Here, a learning algorithm needs to be selected first from the following candidates:
approximate $k$ nearest neighbours~\cite{hnsw}, elastic net linear models~\cite{friedman10glmnet}, random forests~\cite{wright17ranger}, decision trees~\cite{rpart}, support vector machines~\cite{boser92svm}, and gradient boosting~\cite{chen16xgboost}. The hyperparameters of each learner are chosen conditioned on this learner being active, i.e.\ there are hierarchical hyperparameter dependencies. The fidelity of a single evaluation can be controlled by choosing the size of the training data set that is used to train the respective learner. The automated optimization of the pipeline is performed for $89$ different classification tasks~\cite{binder2020}, again taken from OpenML. As a target metric, we opt for the log loss.

\textbf{Neural architecture search}: As third problem scenario we consider is neural architecture search.
The search space of architectures is given by the darts search space \cite{liu19} and architectures were trained and evaluated on CIFAR-10~\cite{cifar}:
A convolutional neural network is constructed by stacking so-called normal and reduction cells that each can be represented as a directed acyclic graph consisting of an ordered sequence of vertices (nodes) resembling feature maps and each directed edge being associated with an operation that transforms the input node.
A tabular representation can be derived using $34$ categorical parameters with $24$ dependencies.
Each architecture can be trained for $1$ to $98$ epochs, allowing again for lower fidelity evaluations.
The target metric is validation accuracy.


\begin{table*}[htbp]
\caption{Three Benchmark Collections of YAHPO Gym used in our Benchmark.}
\centering
\ifarxiv
\scalebox{0.8}{
\fi
\begin{tabular}{llrrrrrrr}
\toprule
& & & \multicolumn{3}{c}{Hyperparameter Types} & & \\ \cmidrule{5-7}
Scenario & Target Metric & $d$ & Cont. & Integer & Categ. & Hierarchical & \# Instances & \# Training Set \\
\midrule 
\textit{lcbench}: HPO of a neural network & cross entropy loss & $7$ & $6$ & $1$ & $0$ & \ding{55} & $35$ & 8 \\
\textit{rbv2\_super}: AutoML pipeline configuration & log loss & $38$ & $20$ & $11$ & $7$ & $\checkmark$ & $89$ & 30\\
\textit{nb301}: Neural architecture search & validation accuracy & $34$ & $0$ & $0$ & $34$ & $\checkmark$ & $1$ & --- \\
\bottomrule
\end{tabular}
\ifarxiv
}
\fi
\label{table:benchmark_instances}
\end{table*}

\subsection{Algorithm Design via Optimization}
We perform experiments both to meta-optimize Alg.~\ref{alg:smashy} and to compare the resulting configuration to the performance of other (MF-)HPO algorithms. Meta-optimization is done on the provided training subset of instances, and performance comparisons are done using all other instances. To investigate to what degree optimal configurations differ between different scenarios, we run optimization for \emph{lcbench} and \emph{rbv2\_super} separately and compare the results. There is only one instance of the \emph{nb301} optimization task, which is therefore exclusively used as a test set and not for optimization.

To meta-optimize the given algorithm, and to draw conclusions about its performance in comparison with other algorithms, a budget endpoint needs to be chosen at which optimization performance is measured. We chose a budget of $30 \cdot d$, with $d$ the dimension of the problem instance. Algorithm configurations are evaluated by running Alg.~\ref{alg:smashy} on a benchmark instance until this budget is exhausted, and the best value reached until then is used as overall outcome.

Each problem instance is evaluated once for each configuration evaluation and the average loss value across instances is used to approximate the expected value in \eqref{eq:ac_objective}. The number of problem instances is relatively small, and we chose to evaluate performance by running surrogate optimization in parallel on a single machine, as opposed to using more sophisticated racing approaches that could potentially save some evaluations. Meta-optimization is done via Bayesian optimization, using the LCB infill criterion and interleaved random meta-configurations every 3 evaluations.

The meta-optimization is performed three times for each scenario, and the resulting performance values are combined into a single dataset to get the overall resulting configuration.
This results in two configurations for our optimizer, $\bm{\gamma}^{*\textit{lcbench}}$ and $\bm{\gamma}^{*\textit{rbv2\_super}}$ optimized on training problems in \textit{lcbench} and \textit{rbv2\_super} respectively.

\subsubsection{Optimization Search Space}
The search space of configurations for Alg.~\ref{alg:smashy} used is shown in Appendix~\ref{app:searchspace}, Table~\ref{tbl:searchspace}. 

While the batch size $\mu$ is constant in the \texttt{equal} \textit{batch\_method}, it is changing for every bracket when \textit{batch\_method} is \texttt{HB}. The batch sizes of $\mu(2), \mu(3), \ldots$ are constructed from $\mu(1)$ dynamically such that the total budget spent on each batch is approximately constant, similar to Hyperband's method~\cite{li2017hyperband} but extended to the cases where $\eta_{\text{surv}} \neq \eta_{\text{fid}}$. We try several surrogate learners: Random forests~\cite{breimanrandomforest} (\texttt{RF}), K-nearest-neighbors with $k$ set to 1 (\texttt{KNN1}), kernelized K-nearest-neighbors with ``optimal'' weighting~\cite{Samworth2012} (\texttt{KKNN7}), and the ratio of density predictions of good and bad points, similar to tree parzen estimators~\cite{bergstra11treeparzen} without a hierarchical structure as done in BOHB~\cite{falkner18bohb} (\texttt{TPE}). For the pre-filtering sample distribution $\P_{\lambdav}(\archive)$, we try both uniform sampling (\texttt{uniform}), and sampling from the estimated density of good points, as done in BOHB~\cite{falkner18bohb} (\texttt{KDE}). $\rho_\text{random}$ determines whether the surrogate model makes predictions assuming the highest fidelity value $r$ observed (\texttt{TRUE}), as opposed to assuming the fidelity of the points being sampled; in the framework of Algs. \ref{alg:tournament} and~\ref{alg:progressive}, this relates to preprocessing done by $\inducer_{f_\text{surr}}$. Note that the maximum number of fidelity steps per batch $s$ is not part of the searchspace: Instead, it is inferred automatically from $\eta_{\text{fid}}$ and the ratio of the maximal and minimal bounds set for $r$ as part of the optimization problem.

\subsection{Algorithm Analysis}
Our goal in this work is not only to determine a configuration of Alg.~\ref{alg:smashy} that performs well on the given HPO problems, but also to determine what effect individual components have on performance. 
To achieve this, we ran experiments on the ``test'' instances provided by YAHPO Gym, both with configurations of our algorithm, and with configurations of our algorithm that re-implement popular, published methods (c.f.\ Table \ref{tab:algorithms}). 
Every evaluation, i.e., a complete HPO run on a problem instance, is repeated $30$ times (with different seeds) to allow for statistical analysis.

We aim to answer the following research questions: 
\begin{description}
    \item[RQ1:] How does the optimal configuration differ between problem scenarios, i.e., do different problem scenarios benefit from different HPO algorithms? 
    \item[RQ2:] How does the optimized algorithm compare to other established HPO algorithms?
    \item[RQ3:] Does the successive-halving fidelity schedule have an advantage over the (simpler) equal-batch-size schedule?
    \item[RQ4:] What is the effect of using multi-fidelity methods in general?
    \item[RQ5a:] Does changing \textsc{Sample} configuration parameters throughout the optimization process give an advantage?
    \item[RQ5b:] Does (more complicated) surrogate-assisted sampling in \textsc{Sample} provide an advantage over using simple random sampling?
    \item[RQ6:] What effect do different surrogate models (or using no model at all) have on the final performance?


\end{description}

A simple method to answer many of these questions is to take the optimized configuration of Alg.~\ref{alg:smashy} and swap out components of it for generally simpler components, thereby performing a one-factor-at-a-time analysis or an ablation study. However, the optimal values of some components may have a strong influence on the choice of other components. We therefore run separate optimizations that are constrained in some ways to reflect some of the research questions. See Table~\ref{tbl:rqtblb} for the different values of $\bm{\gamma}$ we generate. For each value of $\bm{\gamma}$, we run the respectively configured HPO algorithm on both the \emph{lcbench} and the \emph{rbv2\_super} setting, and, unless stated otherwise, once each for \emph{batch\_method} being \texttt{equal} and \texttt{HB}. We refer to an optimized configuration that was obtained on the \emph{lcbench} scenario with \emph{batch\_method} set to \texttt{equal} as $\bm{\gamma}^{*\textit{lcbench}}[\texttt{equal}]$, and to the overall optimum (i.e.\ the better of $\bm{\gamma}^{*\textit{lcbench}}[\texttt{equal}]$ and $\bm{\gamma}^{*\textit{lcbench}}[\texttt{HB}]$) as $\bm{\gamma}^{*\textit{lcbench}}$; similar for \emph{rbv2\_super}. In the following, we describe each question, and how we intend to answer it, in more detail.

\vspace{\baselineskip}
\textbf{RQ1 -- Variation between scenarios:}
We investigate the difference in the values that $\bm{\gamma}^{*\textit{lcbench}}$ and $\bm{\gamma}^{*\textit{rbv2\_super}}$ take, and how large these differences are compared to the natural uncertainty of the optimization outcome. We also investigate how their performance behaves. This gives an idea how robust the results are, and how reliably they perform in new environments.

\textbf{RQ2 -- Optimized algorithm compared to other algorithms:}
We evaluate several well-known HPO algorithms in their default-configuration on the same benchmark instances: for BOHB~\cite{falkner18bohb} we use the implementation found in \texttt{HpBandSter}\footnote{\url{https://github.com/automl/HpBandSter}} version 0.7.4, for HB~\cite{li2017hyperband} we use \texttt{mlr3hyperband}\footnote{\url{https://cran.r-project.org/package=mlr3hyperband}} version 0.1.2, for SMAC~\cite{hutter11smac} we use the SMACv3 package\footnote{\url{https://github.com/automl/SMAC3}} version 1.0.1. SMAC does Bayesian optimization with a random forest model, so we included mlrMBO~\footnote{\url{https://cran.r-project.org/package=mlrMBO}} version 1.1.5 for Gaussian process based BO (GPBO)~\cite{jones98ego}; however, mlrMBO only uses a GP for numerical search spaces and defaults to a random forest for categorical spaces, so we only evaluate GPBO on \emph{lcbench}.
Note that GPBO, SMAC and RS are not multi-fidelity algorithms and therefore always evaluate points with maximum fidelity 1.

\textbf{RQ3 -- Advantage of SH fidelity scheduling:}
It is likely that the relationship between many configuration parameters and algorithm performance depends to a large degree on whether \texttt{equal} or \texttt{HB} fidelity scheduling is used. We therefore perform the overall optimization with both settings separately, and compare their resulting performances.

\textbf{RQ4 -- Effect of using multi-fidelity methods:}
By setting $\eta_\mathrm{fid}$ to $\infty$, and therefore the number of fidelity stages $s$ to 1, we ensure that only evaluations with maximum fidelity 1 are evaluated.

\textbf{RQ5a and RQ5b -- Changing configuration parameters during optimization, and within \textsc{Sample} evaluations:}
We optimized $\gamma_2$, where all configuration parameters that could otherwise depend on $t$ were restricted to being constant, i.e. to have their beginning and end point be equal (RQ5a). In addition to this, we ran another optimization where we further restricted $N_\text{s}^0$ and $N_\text{s}^1$ to be equal, $n_\mathrm{trn}$ to be 1, and only the \texttt{tournament} \textit{filter\_method} be used (RQ5b). The ``test'' performance of the resulting configurations gives an indication for the performance that is lost for the gain in simplicity.

\textbf{RQ6 -- Effect of surrogate models:}
We evaluate the overall result $\bm{\gamma}^*[\texttt{equal}]$ with $\inducer_{f_\text{sur}}$ set to each of the inducers in the original search space (see Tbl.~\ref{tab:smashyspace}). Furthermore, $\bm{\gamma}^*[\texttt{equal}]$ is evaluated with $\rho$ set to 1 (all points are randomly sampled, independent of the filtering model), and finally, with $\rho=1$ and $\P_{\lambdav}(\archive)=\texttt{uniform}$ (all points are sampled completely model-free). 

\textbf{RQ7 -- Performance with parallel resources:}
As optimization of expensive machine learning methods is often run in parallel settings, we evaluate the performance of our method, as well as other methods, in a (simulated) parallel setting. We evaluate our $\lambda^*[\texttt{equal}]$ with $\mu$ set to 32, and with an optimization budget of $30 \cdot 4 \cdot d$, where $d$ is the dimension of the optimization problem. We compare it to GPBO with qLCB~\cite{hutter_parallel_2012} with 32 objective evaluations in parallel and we simulate parallel execution of RS by running enough random evaluations to fill up $30 \cdot 4 \cdot d$ evaluations. Both BOHB and SMAC offer parallelized versions, but the YAHPO Gym benchmark package does not yet provide support for asynchronous parallel evaluation~\cite{pfisterer_yahpo_2021}. However, since HB and BOHB propose evaluations in batches, we compared HB and BOHB by accounting for submitted batches of objective evaluations in increments of 32, essentially simulating a single HB/BOHB optimizer sending objective evaluation tasks to 32 parallel workers and waiting for their completion in a synchronous fashion. 

\subsection{Reproducibility and Open Science} 

The implementation of the framework of Alg.~\ref{alg:smashy} as well as reproducible scripts for the algorithm configuration and analysis are made available via public repositories\footnote{\url{https://github.com/mlr-org/miesmuschel/tree/smashy_ex}, \\ \url{https://github.com/compstat-lmu/paper_2021_benchmarking_special_issue}}. All data that was generated by those analyses is made available as well.

\begin{table*}[h]
\centering
\caption{Summary of Experiment. Shown are the various optimizer configurations $\bm{\gamma}$ that were obtained for our experiments. ``Name'': The name by which we refer to the configuration in the text. ``RQ'': The research question that mainly relates to the configuration. ``Optimize'': Whether the given configuration was obtained by conducting a (possibly constrained) optimization (\checkmark{}), or by substituting values into the global optimum $\bm{\gamma}^*$.}\label{tbl:rqtblb}
\begin{tabular}{lll|l}
\toprule
Signifier         & RQ      & Optimize & Design Modification                                                                                                                                                                                        \\ \midrule
$\bm{\gamma}^*$ & 1, 2, 3 & \checkmark  & none (global optimization)                                                                                                                                                                                 \\
$\bm{\gamma}_1$ & 4       & \ding{55} & $\eta_{\mathrm{fid}}\rightarrow \infty$                                                                                                                                                                 \\
$\bm{\gamma}_2$ & 5a      & \checkmark  & $n_\mathrm{trn}(0)=n_\mathrm{trn}(1),~N_\text{s}^0(0) = N_\text{s}^0(1),~N_\text{s}^1(0) = N_\text{s}^1(1),~\rho(0) = \rho(1)$                                                  \\
$\bm{\gamma}_3$ & 5b      & \checkmark  & \textit{filter\_method}$\rightarrow$ \texttt{tournament}, $n_\mathrm{trn}\rightarrow 1,~N_\text{s}^0(0) = N_\text{s}^0(1) = N_\text{s}^1(0) = N_\text{s}^1(1),~\rho(0) = \rho(1)$ \\
$\bm{\gamma}_4$ & 6       & \ding{55} & \textit{batch\_method}$\rightarrow$ \texttt{equal}, $\inducer_{f_\text{sur}}\rightarrow *$                                                                                                           \\
$\bm{\gamma}_5$ & 6       & \ding{55} & \textit{batch\_method}$\rightarrow$ \texttt{equal}, $\rho\rightarrow 0$                                                                                                                              \\
$\bm{\gamma}_6$ & 6       & \ding{55} & \textit{batch\_method}$\rightarrow$ \texttt{equal}, $\rho\rightarrow 0,~\P_{\lambdav}(\archive)\rightarrow \texttt{uniform}$                                                                                                                              \\
$\bm{\gamma}_7$ & 7       & \ding{55} & \textit{batch\_method}$\rightarrow$ \texttt{equal}, $\mu\rightarrow 32$, quadruple budget \\
\bottomrule                                   
\end{tabular}

\end{table*}

%% file: 50_results.tex
\subsection{General Results}
\begin{figure*}[ht]
    \centering
    \includegraphics[width = 0.7\linewidth]{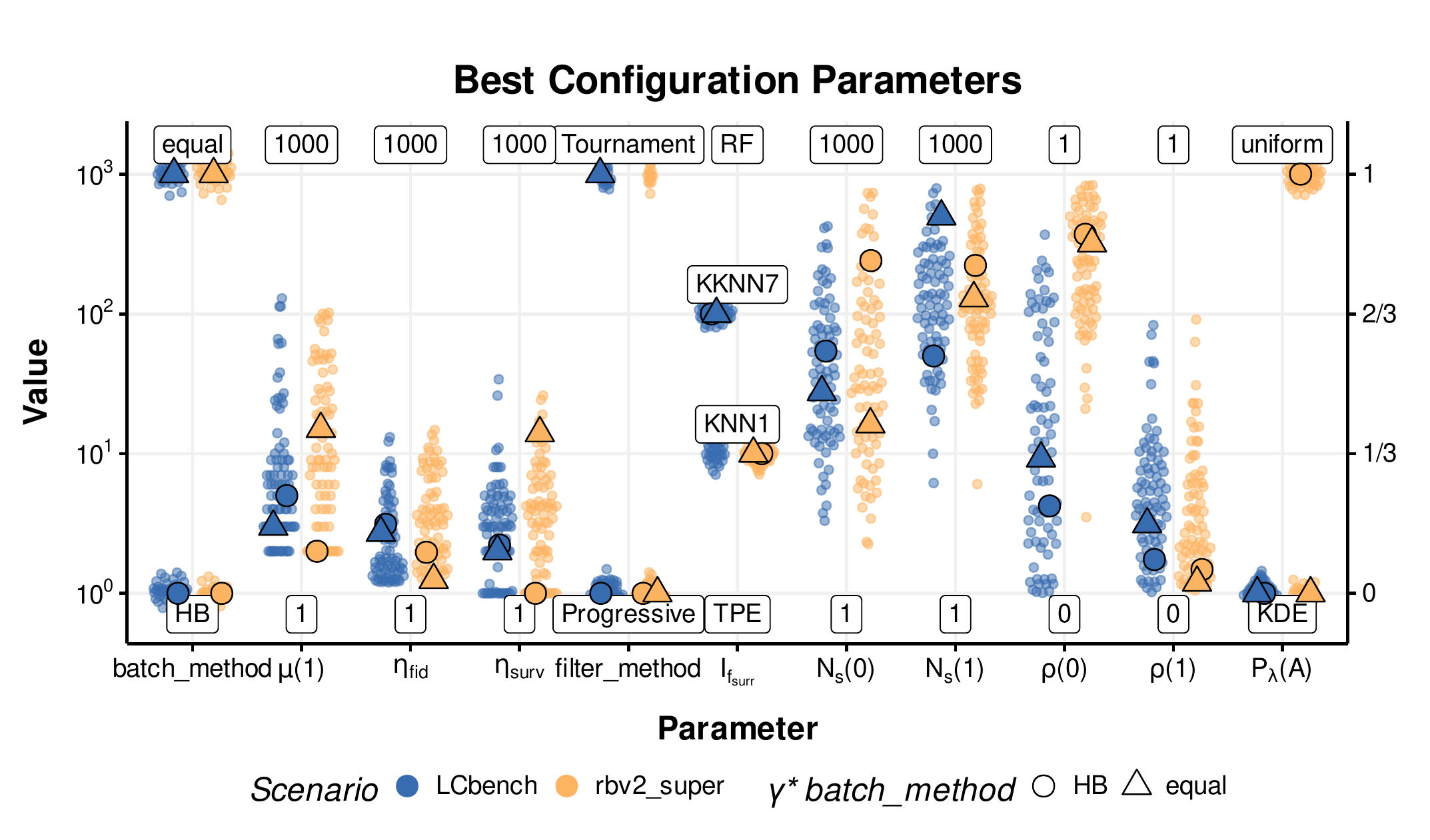}
    \caption{Beeswarm plot of best configurations according to surrogate model over meta-optimization archive of $\gamma^*$. Shown are the top 80 configuration points, according to surrogate model predicted performance, that were evaluated during optimization. Discrete parameters have their levels shown. Most numeric parameters are on a log-scale (left axis), except for $\rho(0)$, $\rho(1)$, which are on a linear scale (right axis). Instead of showing both $N_\text{s}^0(t)$ and $N_\text{s}^0(t)$, their geometric mean, named $N_\text{s}(t)$, is shown. The highlighted large points are $\gamma^*[\texttt{HB}]$ and $\gamma^*[\texttt{EQUAL}]$ that were found on both benchmark scenarios.}
    \label{fig:coordinateplot}
\end{figure*}

\begin{figure*}[ht]
    \centering
    \includegraphics[width = 0.8\linewidth]{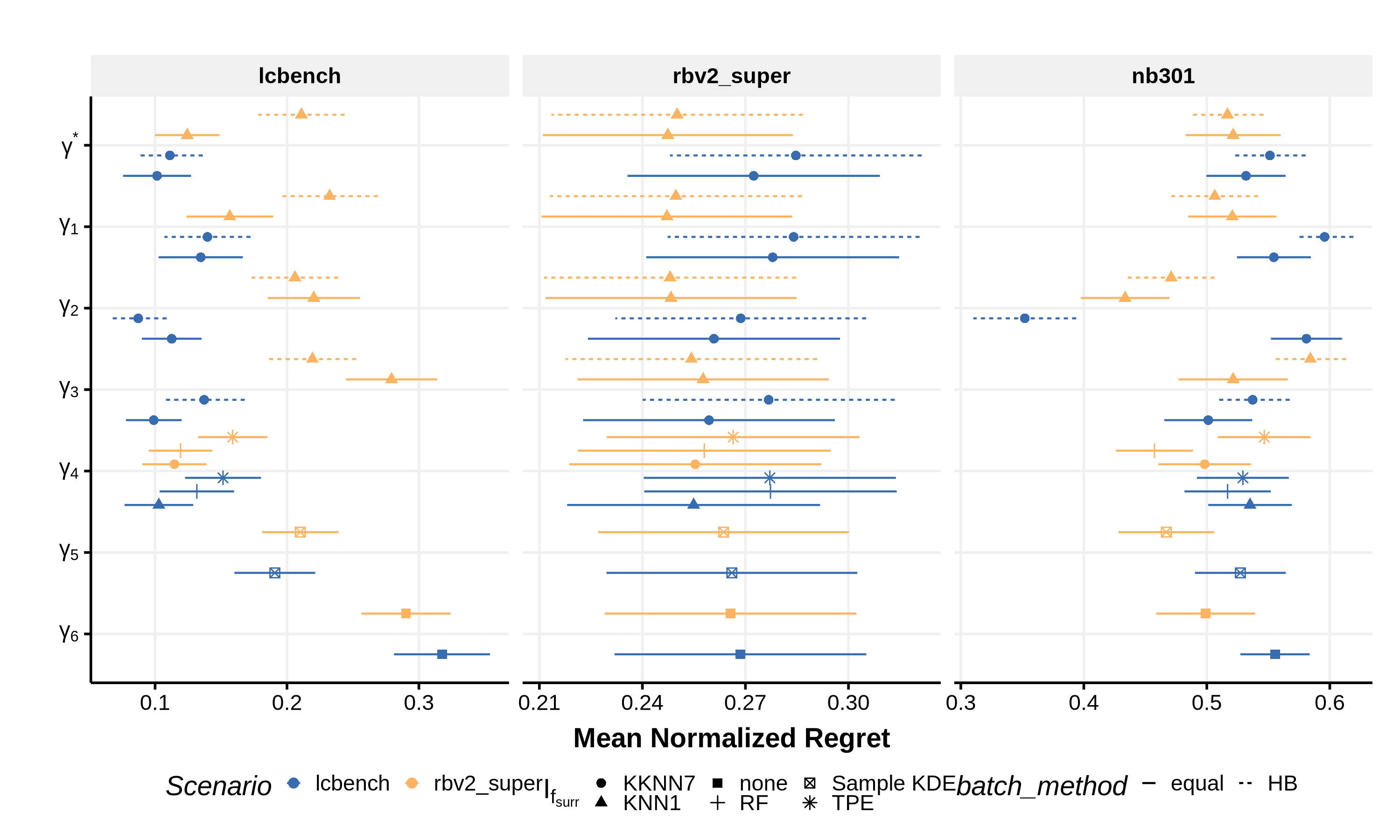}
    \caption{Mean normalized regret of final performance on ``test'' benchmark instances for the configuration shown in Table~\ref{tbl:rqtblb}. Shown is the mean over 30 evaluations, averaged over all available test benchmark instances for each of the three scenarios, the uncertainty bands show standard error over instance means. Regret is calculated as the difference from the so far best evaluation performance to the overall smallest value found on each benchmark instance throughout all experiments, and then normalized so that the value 1 coincides with the median of the performance of all randomly sampled full fidelity evaluations.}
    \label{fig:ablation}
\end{figure*}

Table~\ref{tbl:gammaresults} in Appendix~\ref{app:gammaresults} shows the configuration parameters that were selected, for each benchmark scenario, search space restriction, and for both \textit{batch\_method} variants. 
For both benchmark scenarios the configurations with the $\texttt{HB}$ batch method, i.e., $\gamma^*[\texttt{HB}]$, were returned as best on the training instances, although the difference is small. Figure~\ref{fig:coordinateplot} shows the configuration values of the top 80 evaluated points according to their surrogate predicted performance. The implied ranges covered by the bee swarms are again an indicator of approximate ranges of configurations that can be expected to work well. Figure~\ref{fig:ablation} shows the final performance at $30 \cdot d$ full-budget evaluations of all optimization runs that were performed. The standard error shown for \emph{lcbench} and \emph{rbv2\_super} reflects the variation between the calculated benchmark-instance-wise performance means, representing uncertainty about the ``true'' performance mean if an infinite number of benchmark instances of the given class of problems were available.


\textbf{RQ1 -- Variation between scenarios:}
As can be seen in Table~\ref{tbl:gammaresults} and in Figure~\ref{fig:coordinateplot}, many of the selected components of the $\bm{\gamma}^*$ are relatively close to each other across the two scenarios on which they were optimized. It is therefore interesting to note the differences: $\inducer_{f_\text{surr}}$ is restricted to $\texttt{KNN1}$ on \textit{rbv2\_super}, but can also use $\texttt{KKNN7}$ on \textit{lcbench}, which in fact seems to be slightly preferred. $\P_{\lambdav}(\archive)$ may take any of the two values for \textit{rbv2\_super} but is restricted to \texttt{KDE} in \textit{lcbench}. Finally, $\rho(0)$ is close to 1 in the beginning on \textit{rbv2\_super}, and closer to 0 (although still greater than $\rho(1)$) for \textit{lcbench}.

The degree to which the differences in $\bm{\gamma}^*$ make a difference can be observed in Figure~\ref{fig:ablation}. The optimized results seem to perform well in the test sets of the same scenarios as they were configured on. This is although the optimization has happened on a set of benchmark instances distinct from the test instances. It can therefore be said that we are observing an effect of a latent property of the benchmark scenarios, such as their dimensionality or their landscapes. Comparing the absolute outcome values and their variance, however, the advantage is relatively modest.


%

\begin{figure*}[ht]
    \centering
    \ifarxiv
    \includegraphics[width = \linewidth,clip,trim={13cm 13.5cm 13cm 0cm}]{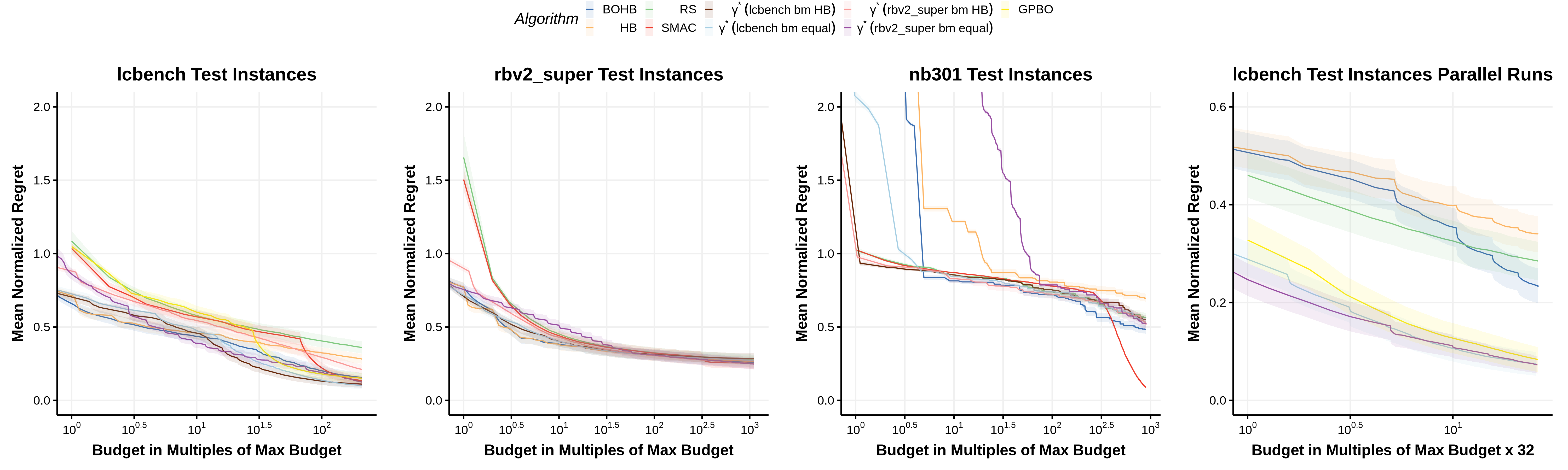}
    \includegraphics[width = \linewidth,clip,trim={0 0 25.5cm 3cm}]{figures/rq1.png}
    \includegraphics[width = \linewidth,clip,trim={25cm 0 0 3cm}]{figures/rq1.png}
    \else
    \includegraphics[width = \linewidth]{figures/rq1.png}
    \fi
    \caption{Optimization progress (mean normalized regret) of serial evaluation on each benchmark scenario (``test'' instances only), as well as 32x parallel evaluation on \emph{lcbench}. ``$\gamma^*$(lcbench~bm~equal)'' is the configuration obtained from optimizing on \emph{lcbench} with \emph{batch\_method} \texttt{equal} etc.  Shown is the mean over 30 evaluations, averaged over all available test benchmark instances for each of the three scenarios, the uncertainty bands show standard error over test instances. Note log scale on x axis. Regret is calculated as the difference from the so far best evaluation performance to the overall smallest value found on each benchmark instance throughout all experiments, and then normalized so that the value 1 coincides with the median of the performance of all randomly sampled full fidelity evaluations. Performance is calculated as the best performance observed by the optimizer and therefore depends on evaluation fidelity; this is the reason for the initially ``slow'' convergence of algorithms that make their first full-fidelity evaluation late. Note on the parallel plot that $\mu$ of $\gamma^*[\texttt{equal}]$ was set to 32 and HB and BOHB were only naively parallelized to simulate a synchronous ``single optimizer, multiple workers'' environment.}
    \label{fig:rq1}
\end{figure*}

\begin{figure*}
     \centering
     \begin{subfigure}[b]{0.45\linewidth}
         \centering
         \includegraphics[width=\linewidth]{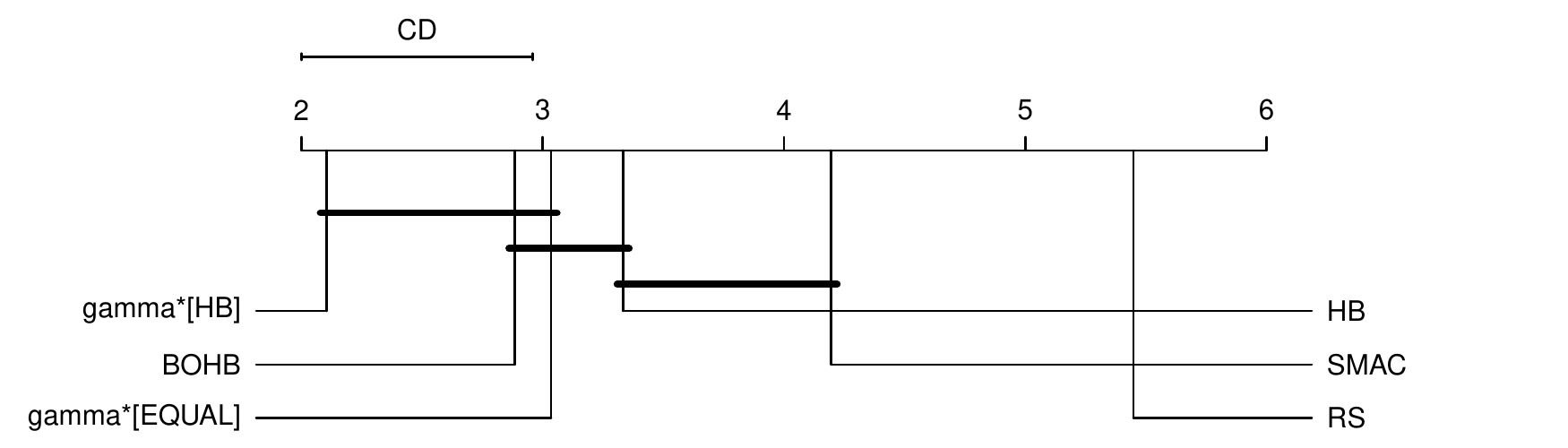}
         \caption{Intermediate optimization budget of $100$ full evaluations}
         \label{fig:cd_intermediate_RQ2_RQ3_RQ4}
     \end{subfigure}
     \hfill
     \begin{subfigure}[b]{0.45\linewidth}
         \centering
         \includegraphics[width=\linewidth]{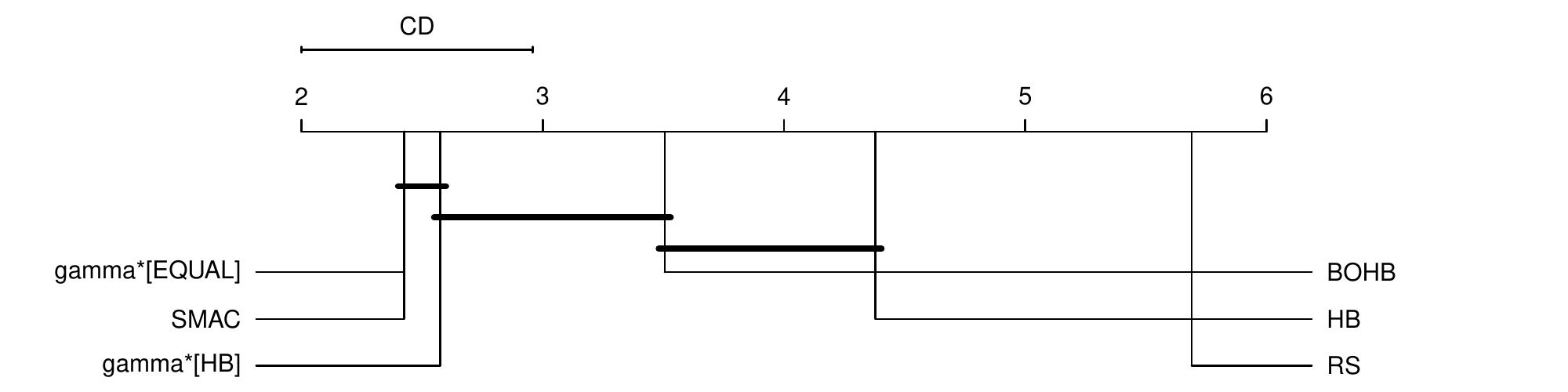}
         \caption{Full evaluation budget (final performance)} 
         \label{fig:cd_final_RQ2_RQ3_RQ4}
     \end{subfigure}
    \caption{Critical difference plot \cite{demsar06cd} to compare the performance of multiple algorithms across all instances and scenarios. Lower ranks are better. Horizontal bold bars indicate that there is no significant difference between algorithms ($\alpha = 1\%$). GPBO, which was not evaluated on all scenarios, is not included.}
    \label{fig:cd_RQ2_RQ3_RQ4}
\end{figure*}

\textbf{RQ2 -- Optimized algorithm compared to other algorithms:}
The performance curves for the mean normalized regret is shown in Figure~\ref{fig:rq1}, and the final performance values at $30\cdot d$ full fidelity evaluations is shown in Figure~\ref{fig:ablation}. A critical difference plot and test can be seen in Figure~\ref{fig:cd_final_RQ2_RQ3_RQ4}. The behavior of RS, HB, BOHB and SMAC is not surprising: Initially RS and SMAC perform the same, because SMAC evaluates an initial random design, after which its performance improves fast. HB and BOHB initially perform the same, and perform better than RS or SMAC because of their multi-fidelity evaluations. After a while, BOHB starts to outperform HB because of its surrogate-based sampling. Therefore, BOHB performs well for most budgets, often being the best optimizer for a budget of 1 full evaluation and for 100 full evaluations.
Given its multifidelity characteristics, Hyperband (HB) is a good choice for low budgets, while SMAC is well suited for larger optimization budgets. 

Our algorithm is very competitive on both \textit{lcbench} and \textit{rbv2\_super}, but gets outperformed by SMAC on \textit{nb301}. We assume this is because the search space of \textit{nb301} is a purely categorical search space with many components, for which SMACs random forest based surrogate model appears to give it an advantage. 

Although our algorithm was only optimized for its performance at $30\cdot d$ evaluations, it still has strong performance after fewer evaluations, as seen in Figure~\ref{fig:cd_final_RQ2_RQ3_RQ4}.

\textbf{RQ3 -- Advantage of SH fidelity scheduling:} In both scenarios, the batch method \texttt{HB} is ultimately selected for the optimum $\gamma^*$, although Figures \ref{fig:cd_final_RQ2_RQ3_RQ4} and~\ref{fig:cd_final_RQ2_RQ3_RQ4} show that the difference to batch size \texttt{equal} is not statistically significant at $\alpha=1\%$. We observe that the \texttt{equal} fidelity scheduling mode has several advantages: For one, it is much simpler than \texttt{HB}, as it does not need to keep track of SH-brackets, and does not need to adapt $\mu(b)$ to make the budget expense at each bracket approximately equal. As another benefit, it allows for easy parallel scheduling of evaluations. This is because it always schedules the same number of function evaluations at a time, which can therefore be done synchronously.

\textbf{RQ4 -- Effect of using multi-fidelity methods:} Our results support the superiority of multi-fidelity HPO methods as compared to HPO methods that do not make use of lower fidelity approximations. 
Fig.~\ref{fig:cd_intermediate_RQ2_RQ3_RQ4} suggests that multi-fidelity methods are significantly better than their non multi-fidelity counterparts if optimization is stopped at an intermediate overall budget corresponding to $100$ full model evaluations. To be more precise, it can be seen that BOHB as well as both optimized variants $\bm{\gamma}^*[\texttt{equal}]$ and $\bm{\gamma}^*[\texttt{HB}]$ (optimized for the respective scenario, respectively) significantly outperform SMAC under this strict budget constraint. In line with \cite{li2017hyperband}, HB significantly outperforms random search under this budget. On the other hand, Fig.~\ref{fig:cd_final_RQ2_RQ3_RQ4} provides evidence that multifidelity methods can achieve final performance on the same level as the state-of-the-art method that is not making use of low fidelity approximations (SMAC).
It can be concluded that the effect of a properly designed multifidelity mechanism provides significant improvements on anytime performance, while not lowering, but potentially also not raising final performance significantly. In our opinion, the gain in anytime performance justifies the additional algorithmic complexity that is introduced by using multi-fidelity methods. 

\textbf{RQ5a, RQ5b -- Changing configuration parameters during optimization, and within \textsc{Sample} evaluations:}
The observations made for $\gamma_2$ (forbidding change over time) and $\gamma_3$ (forbidding change over time and within each batch) are slightly contradictory. In particular, the \emph{nb301} performance of $\gamma_2^{\textit{lcbench}}[\texttt{HB}]$ is a visible outlier with regard to optimization performance. There is no obvious explanation available from inspecting the configuration parameters of $\gamma_2^{\textit{lcbench}}[\texttt{HB}]$, but it is possible that it is an accidental ``good fit'' of configuration parameters to the specific landscape of \emph{nb301}.

On \emph{lcbench} and \emph{rbv2\_super}, the impact of restricting the search space is less and within the possible bounds of random error. We note, however, that both changing configuration parameters over time and within each batch sample introduces significant complexity to the algorithm, therefore making the restricted optimization results obvious favorites over $\gamma^*$.

\textbf{RQ6 -- Effect of surrogate models:}
Surprisingly, the simple k-nearest-neighbors algorithm seems to be chosen systematically by the algorithm configuration for both \emph{lcbench} and \emph{rbv2\_super} (see Fig.~\ref{fig:coordinateplot}), either with a value of $k = 1$ or $k = 7$. Our ablation experiments suggest that the performance of the optimizer is on average best when using this surrogate learner, even though the differences do not seem to be significant. Under the aspect of algorithmic complexity, it seems that the \texttt{KNN1} is a reasonable alternative to more complex surrogate learners like the \texttt{TPE}-based method proposed by the original BOHB algorithm.

\textbf{RQ7 -- Performance with parallel resources:}
The results shown in Fig.~\ref{fig:rq1} show that our algorithm is competitive with GPBO, a state-of-the-art synchronously parallel optimization algorithm, when evaluated with 32x parallel resources. This result also shows the main advantage that the \texttt{equal} fidelity schedule has over scheduling like \texttt{HB}: synchronously parallelizing HB or BOHB puts them at a great disadvantage over even RS; for these it is necessary to use asynchronously parallelized methods~\cite{klein20abohb,li20asha} or use an archive shared between multiple workers~\cite{falkner18bohb} to get competitive results. 
However, synchronous objective evaluations are much easier to implement in many environments than asynchronous communication between workers, making the advantage of the simplicity of the \texttt{equal} schedule even more pronounced.

%% file: 90_conclusion.tex

We presented a principled approach to benchmark-driven algorithm design by applying it to generic multi-fidelity HPO.
For this, we formalized the search space for our rich and configurable optimization framework.
Given the search space, we applied Bayesian optimization for meta-optimization of our framework on two different problem scenarios within the field of AutoML. The configured algorithms are evaluated and compared to BOHB, Hyperband, SMAC, and a simple random search as reference. We performed an extensive analysis to evaluate the effect of algorithmic components on performance against the additional algorithmic complexity they introduce, observing that we achieve decent performance compared to widely used HPO-algorithms.

We conclude that the additional algorithmic complexity introduced by handling of multi-fidelity evaluations provides significant benefits. However, based on our experiments we argue that design choices made by established multifidelity optimizers like BOHB can be replaced by simpler choices: For example, the (more complex) successive halving schedule is not significantly better than a schedule using equal batch sizes. Switching to equal batch sizes can simplify parallelization drastically by offering the opportunity of efficient synchronous parallelization while staying competitive with other parallelized optimization methods.

An aspect of BOHB that was consistently chosen by the optimizer to be included in the search space is the kernel-density estimator-based sampling of new proposed points, whether filtered by a surrogate model or not. This detail, which is not usually presented as the main feature of BOHB, seems to have an unexpectedly large impact. On the other hand our optimization results suggest a surprisingly simple surrogate learner (\texttt{knn}, $k=1$), even outperforming the choices made in BOHB.

Some components of our search space with potential algorithmic complexity have not shown much benefit. Optimization on \emph{rbv2\_super} did choose time-varying random interleaving, and overall there was a slight favoring of more agressive filtering late during an optimization run ($N_{\text{s}}(1) > N_{\text{s}}(0)$), but the results did not consistently outperform a configuration obtained from a restricted optimization that excluded time-varying configuration parameters.

Our analysis of the set of best observed performances during optimization indicates there is a large agreement between benchmark scenarios about what the optimal $\gamma^*$ configuration should be, the most notable distinction being parameters that control (model-based) sampling and the surrogate model. This suggests there may be a set of configuration parameters that are either generally good in many machine machine learning problems, or do have little influence on performance and can therefore be set to the simplest value. There seems to be a remaining set of configuration parameters that should be adapted to the properties of the optimization problem. The meta-optimization framework presented in this work could be used in future work to investigate the relationship between features of optimization problems and related optimal configuration. The knowledge gained could ultimately used for exploratory landscape analysis to set configuration parameters automatically~\cite{mersmann2011exploratory}.

Other future work indicated because of some of the drawbacks of the benchmarks used: We were not able to optimize, or compare to, methods that use asynchronous proposals of evaluations. This is a major drawback, because asynchronous methods are important in the current age where parallel resources are plentiful, but current widely-used surrogate-based benchmarks do not allow for the easy evaluation of asynchronous optimization performance. Suggested methods, such as waiting with a sleep-timer for an appropriate amount~\cite{falkner18bohb}, are impractical for meta-optimization. A method based on synchronizing objective evaluation return times by the benchmark suite has been suggested~\cite{pfisterer_yahpo_2021} and, when available, would make optimization of asynchronous algorithms possible.

%% file: 95_appendix.tex
\section{Sample Algorithms}\label{app:algs}

The pseudocode of both \textsc{Sample} algorithms is presented here for clarity.

\begin{algorithm}[H]
\caption{\textsc{SampleTournament} algorithm}
\label{alg:tournament}
    \hspace*{\algorithmicindent} \textbf{Input}: Archive $\archive$, number of points to generate $\mu$, current fidelity $r$
    
    \vspace*{0.2cm}
    \hspace*{\algorithmicindent} \textbf{Configuration Parameters}: Surrogate learner $\inducer_{f_\text{sur}}$, generating distribution $\P_{\lambdav}(\archive)$, random interleave fraction $\rho$, sample filtering rates $(N_\text{s}^0, N_\text{s}^1)$, points to sample per tournament round $n_\mathrm{trn}$.
    
    \vspace*{0.2cm}
    \hspace*{\algorithmicindent} \textbf{State Variables}: Batch of proposed configurations $C \leftarrow \emptyset$
\begin{algorithmic}[1]
    \vspace*{0.2cm}
    \State Use $\rho$ to decide how many points $n_\mathrm{random\_interleave}$ to sample without filter
    \State $C \leftarrow$ Sample $n_\mathrm{random\_interleave}$ configurations from $\P_{\lambdav}(\archive)$
    \State $\mu\leftarrow\mu - n_\mathrm{random\_interleave}$
    \State $n \leftarrow \lceil\mu / n_\mathrm{trn}\rceil$\Comment{Numter of tournament rounds}
    \State $f_\text{sur}\leftarrow \inducer_{f_\text{sur}}(\archive)$\Comment{Surrogate model}
    \For{$i \leftarrow 1$ \textbf{to} $n$}
        \State $n_\mathrm{sample}\leftarrow\left\lfloor(N_\text{s}^0)^\frac{n-i}{n-1}\cdot (N_\text{s}^1)^\frac{i-1}{n-1}\right\rceil$
        \State $C_0 \leftarrow$ Sample $n_\mathrm{sample}$ configurations from $\P_{\lambdav}(\archive)$
        \State Predict performances of points in $C_0$ using $f_\text{sur}$
        \State $C \leftarrow C \cup \textsc{select\_top}\left(C_0, \min(n_\mathrm{trn}, \mu - |C|)\right)$
    \EndFor \\
    \Return{C}
    \vspace*{0.2cm}
\end{algorithmic}
\end{algorithm}

\begin{algorithm}[H]
\caption{\textsc{SampleProgressive} algorithm}
\label{alg:progressive}
    \hspace*{\algorithmicindent} \textbf{Input}: Surrogate learner $\inducer_{f_\text{sur}}$, Archive $\archive$, number of points to generate $\mu$, current fidelity $r$, random interleave fraction $\rho$, sample filtering rates $(N_\text{s}^0, N_\text{s}^1)$, generating distribution $\P_{\lambdav}(\archive)$
    
    \vspace*{0.2cm}
    \hspace*{\algorithmicindent} \textbf{State Variables}: Batch of proposed configurations $C \leftarrow \emptyset$, (ordered) pool of sampled points to select from $\mathcal{P}$
\begin{algorithmic}[1]
    \vspace*{0.2cm}
    \State Use $\rho$ to decide how many points $n_\mathrm{random\_interleave}$ to sample without filter
    \State $C \leftarrow$ Sample $n_\mathrm{random\_interleave}$ configurations from $\P_{\lambdav}(\archive)$
    \State $\mu\leftarrow\mu - n_\mathrm{random\_interleave}$
    \State $n_\mathrm{pool} \leftarrow \mu\cdot\textrm{max}(N_\text{s}^0, N_\text{s}^1)$
    \State $\mathcal{P}\leftarrow$ Sample $n_\mathrm{pool}$ configurations from $\P_{\lambdav}(\archive)$
    \State $f_\text{sur}\leftarrow \inducer_{f_\text{sur}}(\archive)$\Comment{Surrogate model}
    \State Predict performances of points in $\mathcal{P}$ using $f_\text{sur}$
    \For{$i \leftarrow 1$ \textbf{to} $\mu$}
        \State $n_\mathrm{options}\leftarrow\left\lfloor(N_\text{s}^0)^\frac{\mu-i}{\mu-1}\cdot (N_\text{s}^1)^\frac{\mu-1}{\mu-1}\right\rceil$
        \State $\mathcal{P}_\mathrm{options} \leftarrow$ first $n_\mathrm{options}$ elements of $\mathcal{P}$ 
        \State $S \leftarrow \textsc{select\_top}\left(\mathcal{P}_\mathrm{options}, 1\right)$
        \State $C \leftarrow C \cup S$
        \State $\mathcal{P} \leftarrow \mathcal{P} - S$
    \EndFor \\
    \Return{C}
    \vspace*{0.2cm}
\end{algorithmic}
\end{algorithm}

\section{Meta-Optimization Search Space}\label{app:searchspace}

The full optimization space used for optimization of $\gamma^*$ is presented here in Table~\ref{tbl:searchspace}; Other $\gamma$ results have the restrictions applied to them as shown in Tbl.~\ref{tbl:rqtblb} in Section~\ref{sec:40_experimental_design}.

\begin{table*}[htbp]
\caption{Meta-optimization search space used to configure Alg.~\ref{alg:smashy}. Some configuration parameters are optimized on a non-linear \textit{scale}, meaning e.g.\ the optimizer optimizes a value of $\log{\mu(1)}$ ranging from $\log{2}$ to $\log{200}$.}\label{tbl:searchspace}
\centering
\ifarxiv
\scalebox{0.8}{
\fi
\begin{tabular}{l l l l}
\toprule
Parameter                       & Meaning                           & Range & Scale \\ \midrule
$\mu(1)$                        & (first bracket) batch size        & $\{2, \ldots, 200\}$ & $\log{\mu(1)}$ \\
\textit{batch\_method}        & batch method                      & \{\texttt{equal}, \texttt{HB}\}\\
$\eta_{\text{fid}}$             & fidelity rate                     & $[2^{1/4}, 2^4]$ & $\log{\log{\eta_{\text{fid}}}}$ \\
$\eta_{\text{surv}}$            & survival rate                     & $[1, \infty)$ & $1/\eta_{\text{surv}}$\\
\textit{filter\_method}       & \textsc{Sample} method            & \{\textsc{SampleTournament}, \textsc{SampleProgressive}\}\\
$\P_{\lambdav}(\archive)$               & \textsc{Sample} generating distribution       & \{\texttt{uniform}, \texttt{KDE}\}\\
$\inducer_{f_\text{surr}}$      & surrogate learner                 & \{\texttt{KNN1}, \texttt{KKNN7}, \texttt{TPE}, \texttt{RF}\}\\
$n_{\text{trn}}(0)$             & filter sample per tournament round at $t=0$ & \{1, \ldots, 10\} & $\log{n_{\text{trn}}(0)}$\\
$n_{\text{trn}}(1)$             & filter sample per tournament round at $t=1$ & \{1, \ldots, 10\} & $\log{n_{\text{trn}}(1)}$\\
$N_\text{s}^0(0)$               & filtering rate of first point in batch at $t=0$ & $[1, 1000]$ & $\log{N_\text{s}^0(1)}$\\
$N_\text{s}^0(1)$               & filtering rate of first point in batch at $t=1$ & $[1, 1000]$ & $\log{N_\text{s}^0(1)}$\\
$N_\text{s}^1(0)$               & filtering rate of last point in batch at $t=0$ & $[1, 1000]$ & $\log{N_\text{s}^1(1)}$\\
$N_\text{s}^1(1)$               & filtering rate of last point in batch at $t=1$ & $[1, 1000]$ & $\log{N_\text{s}^1(1)}$\\
$\rho(0)$                       & random interleave fraction at $t=0$ & $[0, 1]$\\
$\rho(1)$                       & random interleave fraction at $t=1$ & $[0, 1]$\\
\textit{filter\_mb}           & surrogate prediction always with maximum $r$ & \{\texttt{TRUE}, \texttt{FALSE}\}\\
$\rho_\text{random}$            & random interleave the same number in every batch & \{\texttt{TRUE}, \texttt{FALSE}\}\\
\bottomrule
\end{tabular}
\ifarxiv
}
\fi
\label{tab:smashyspace}
\end{table*}

\section{All $\bm{\gamma}$-Values}\label{app:gammaresults}

A table of all optimization $\gamma$-values is given in Tbl.~\ref{tbl:gammaresults}.

\begin{table*}
\caption{Optimized configuration parameters, under some constraints. Top: restricted to \textit{batch\_method} \texttt{HB}, bottom: \texttt{equal}. $\gamma_2$, $\gamma_3$ are further restricted as described in Table~\ref{tbl:rqtblb} in Section~\ref{sec:40_experimental_design}. Shown is the overall result. Square brackets show range (for numeric parameters) or list (for discrete parameters) of values found in individual optimization runs when not aggregated, as a rough indicator of uncertainty. ``(!)'' indicates the parameter was forced to the value by a restriction.}\label{tbl:gammaresults}
\centering
\ifarxiv
\scalebox{0.8}{
\fi

\begin{tabular}{l|llllll}
\toprule
Parameter & $\gamma^*$ & $\gamma^*$ & $\gamma_2$ & $\gamma_2$ & $\gamma_3$ & $\gamma_3$\\

Scenario & lcbench & rbv2\_super & lcbench & rbv2\_super & lcbench & rbv2\_super\\
\midrule
\multicolumn{7}{l}{}\\
\multicolumn{7}{l}{Optimized with \textit{batch\_method} \texttt{HB}:}\\
\midrule
$\mu(1)$ & 5 [5, 23] & 2 [2, 52] & 126 [10, 126] & 8 [8, 114] & 5 [3, 68] & 2 [2, 52]\\\hline
$\eta_{\text{fid}}$ & 3.11 [1.25, 3.11] & 1.97 [1.97, 6.73] & 2.19 [1.68, 10.2] & 4.4 [2.04, 4.4] & 14.6 [1.45, 14.6] & 5.19 [2.24, 5.19]\\\hline
$\eta_{\text{surv}}$ & 2.22 [2.22, 6.1] & 6.09 [1.65, 6.09] & 3.42 [2.58, 9.19] & 3.26 [3.26, 5] & 1.15 [1.15, 3.07] & 1.20 [1.03, 1.62]\\\hline
\textit{filter\_method} & \textsc{Prog} [\textsc{Trn}] & \textsc{Prog} [\textsc{Trn}] & \textsc{Prog} [\textsc{Trn}] & \textsc{Prog} [\textsc{Trn}] & \textsc{Trn} (!) & \textsc{Trn} (!)\\\hline
$\P_{\lambdav}(\archive)$ & \texttt{KDE} & \texttt{uniform} [\texttt{KDE}] & \texttt{KDE} & \texttt{uniform} [\texttt{KDE}] & \texttt{KDE} & \texttt{uniform}\\\hline
$\inducer_{f_\text{surr}}$ & \texttt{KKNN7} [\texttt{KNN1}] & \texttt{KNN1} & \texttt{KKNN7} [\texttt{KNN1}] & \texttt{KNN1} & \texttt{KKNN7} [\texttt{KNN1}] & \texttt{KNN1}\\\hline
$n_{\text{trn}}(0)$ & 2 [2, 8] & 2 [1, 5] & \multirow{2}{*}{5 [1, 8]} & \multirow{2}{*}{5 [1, 6]} & \multirow{2}{*}{1 (!)} & \multirow{2}{*}{1 (!)}\\\cline{1-3}
$n_{\text{trn}}(1)$ & 1 [1, 5] & 5 [1, 5] &  & & & \\\hline
$N_\text{s}^0(0)$ & 101 [9.19, 124] & 226 [2.03, 226] & \multirow{2}{*}{39.6 [10.5, 76.7]} & \multirow{2}{*}{125 [125, 163]} & \multirow{4}{*}{570 [73, 570]} & \multirow{4}{*}{155 [155, 561]}\\\cline{1-3}
$N_\text{s}^0(1)$ & 312 [56.3, 817] & 495 [57.1, 533] &  & & & \\\cline{1-5}
$N_\text{s}^1(0)$ & 28.9 [4.84, 144] & 256 [7.19, 256] & \multirow{2}{*}{31.4 [18.2, 74.6]} & \multirow{2}{*}{481 [480, 563]} &  & \\\cline{1-3}
$N_\text{s}^1(1)$ & 8 [8, 654] & 99.7 [46.4, 890] &  & &   & \\\hline
$\rho(0)$ & 0.21 [0.12, 0.85] & 0.86 [0.68, 0.86] & \multirow{2}{*}{0.37 [0.2, 0.49]} & \multirow{2}{*}{0.71 [0.49, 0.71]} & \multirow{2}{*}{0.38 [0.12, 0.54]} & \multirow{2}{*}{0.34 [0.34, 0.45]}\\\cline{1-3}
$\rho(1)$ & 0.08 [0.08, 0.55] & 0.06 [0.01, 0.25] &  &  &  & \\\hline
\textit{filter\_mb} & TRUE [FALSE] & FALSE [TRUE] & TRUE & TRUE & TRUE [FALSE] & FALSE\\\hline
$\rho_\text{random}$ & FALSE [TRUE] & TRUE [FALSE] & FALSE [TRUE] & TRUE [FALSE] & TRUE & TRUE [FALSE] \\
\bottomrule
\multicolumn{7}{l}{}\\
\multicolumn{7}{l}{Optimized with \textit{batch\_method} \texttt{equal}:}\\
\midrule
$\mu(1)$ & 3 [2, 4] & 15 [11, 15] & 5 [2, 7] & 5 [2, 5] & 2 [2, 6] & 85 [3, 93]\\\hline
$\eta_{\text{fid}}$ & 2.71 [1.77, 12.2] & 1.25 [1.22, 1.43] & 2.63 [2.27, 6.01] & 8 [1.28, 8] & 2.59 [1.46, 2.59] & 2.3 [1.36, 12.7]\\\hline
$\eta_{\text{surv}}$ & 2.5 [1.23, 3.36] & 18.8 [8.74, 18.8] & 1.87 [1.84, 5.49] & 3.45 [3.45, 5.53] & 3.53 [1.29, 4.86] & 6.5 [5.34, 11.3]\\\hline
\textit{filter\_method} & \textsc{Trn} [\textsc{Prog}] & \textsc{Prog} & \textsc{Trn} [\textsc{Prog}] & \textsc{Prog} [\textsc{Trn}] & \textsc{Trn} (!) & \textsc{Trn} (!)\\\hline
$\P_{\lambdav}(\archive)$ & \texttt{KDE} & \texttt{KDE} [\texttt{uniform}] & \texttt{KDE} & \texttt{uniform} [\texttt{KDE}] & \texttt{KDE} & \texttt{uniform} [\texttt{KDE}]\\\hline
$\inducer_{f_\text{surr}}$ & \texttt{KKNN7} [\texttt{KNN1}] & \texttt{KNN1} & \texttt{KNN1} [\texttt{KNN7}] & \texttt{KNN1} & \texttt{KNN1} [\texttt{KNN7}] & \texttt{KNN1}\\\hline
$n_{\text{trn}}(0)$ & 1 [1, 4] & 2 [1, 3] & \multirow{2}{*}{5 [1, 5]} & \multirow{2}{*}{3 [1, 3]} & \multirow{2}{*}{1 (!)} & \multirow{2}{*}{1 (!)} \\\cline{1-3}
$n_{\text{trn}}(1)$ & 2 [1, 2] & 9 [1, 9] &  & & & \\\hline
$N_\text{s}^0(0)$ & 21.5 [1.63, 309] & 39.5 [2.42, 39.5] & \multirow{2}{*}{169 [43.4, 191]} & \multirow{2}{*}{212 [49.5, 212]} & \multirow{4}{*}{81.3 [24.8, 111]} & \multirow{4}{*}{583 [295, 777]}\\\cline{1-3}
$N_\text{s}^0(1)$ & 941 [58.2, 991] & 18.1 [11.5, 408] &  & &  & \\\cline{1-5}
$N_\text{s}^1(0)$ & 35.4 [7.8, 280] & 6.65 [5.43, 391] & \multirow{2}{*}{4.76 [2.34, 273]} & \multirow{2}{*}{1.71 [1.71, 4.21]} &  & \\\cline{1-3}
$N_\text{s}^1(1)$ & 264 [5, 474] & 925 [25.4, 925] &  &  &  & \\\hline
$\rho(0)$ & 0.32 [0.09, 0.68] & 0.83 [0.49, 0.83] & \multirow{2}{*}{0.34 [0.09, 0.37]} & \multirow{2}{*}{0.34 [0.34, 0.53]} & \multirow{2}{*}{0.27 [0.03, 0.27]} & \multirow{2}{*}{0.96 [0.38, 0.96]}\\\cline{1-3}
$\rho(1)$ & 0.16 [0.06, 0.29] & 0.03 [0.03, 0.5] &  &  & & \\\hline
\textit{filter\_mb} & TRUE & TRUE & TRUE & TRUE & TRUE [FALSE] & TRUE\\\hline
$\rho_\text{random}$ & TRUE [FALSE] & FALSE & TRUE [FALSE] & FALSE [TRUE] & TRUE & TRUE [FALSE]\\\bottomrule
\end{tabular}
\ifarxiv
}
\fi
\end{table*}

%% file: ms.bbl
\begin{thebibliography}{10}
\providecommand{\url}[1]{#1}
\csname url@samestyle\endcsname
\providecommand{\newblock}{\relax}
\providecommand{\bibinfo}[2]{#2}
\providecommand{\BIBentrySTDinterwordspacing}{\spaceskip=0pt\relax}
\providecommand{\BIBentryALTinterwordstretchfactor}{4}
\providecommand{\BIBentryALTinterwordspacing}{\spaceskip=\fontdimen2\font plus
\BIBentryALTinterwordstretchfactor\fontdimen3\font minus
  \fontdimen4\font\relax}
\providecommand{\BIBforeignlanguage}[2]{{%
\expandafter\ifx\csname l@#1\endcsname\relax
\typeout{** WARNING: IEEEtran.bst: No hyphenation pattern has been}%
\typeout{** loaded for the language `#1'. Using the pattern for}%
\typeout{** the default language instead.}%
\else
\language=\csname l@#1\endcsname
\fi
#2}}
\providecommand{\BIBdecl}{\relax}
\BIBdecl

\bibitem{bischl21hpo}
B.~Bischl, M.~Binder, M.~Lang, T.~Pielok, J.~Richter, S.~Coors, J.~Thomas,
  T.~Ullmann, M.~Becker, A.~Boulesteix, D.~Deng, and M.~Lindauer,
  ``Hyperparameter optimization: Foundations, algorithms, best practices and
  open challenges,'' \emph{CoRR}, vol. abs/2107.05847, 2021.

\bibitem{kotthoff2019autoweka}
L.~Kotthoff, C.~Thornton, H.~H. Hoos, F.~Hutter, and K.~Leyton-Brown,
  ``Auto-weka: Automatic model selection and hyperparameter optimization in
  weka,'' in \emph{Automated Machine Learning: Methods, Systems, Challenges},
  F.~Hutter, L.~Kotthoff, and J.~Vanschoren, Eds.\hskip 1em plus 0.5em minus
  0.4em\relax Cham: Springer International Publishing, 2019, pp. 81--95.

\bibitem{wolpert_no_1997}
D.~H. Wolpert and W.~G. Macready, ``No free lunch theorems for optimization,''
  \emph{IEEE Transactions on Evolutionary Computation}, vol.~1, no.~1, pp.
  67--82, 1997.

\bibitem{jones98ego}
D.~R. Jones, M.~Schonlau, and W.~J. Welch, ``Efficient global optimization of
  expensive black-box functions,'' \emph{J. Glob. Optim.}, vol.~13, no.~4, pp.
  455--492, 1998.

\bibitem{hutter11smac}
F.~Hutter, H.~H. Hoos, and K.~Leyton-Brown, ``Sequential model-based
  optimization for general algorithm configuration,'' in \emph{Learning and
  Intelligent Optimization}, C.~A.~C. Coello, Ed.\hskip 1em plus 0.5em minus
  0.4em\relax Berlin, Heidelberg: Springer Berlin Heidelberg, 2011, pp.
  507--523.

\bibitem{swersky14freezethaw}
K.~Swersky, J.~Snoek, and R.~P. Adams, ``Freeze-thaw bayesian optimization,''
  \emph{CoRR}, vol. abs/1406.3896, 2014.

\bibitem{bergstra2012randomsearch}
J.~Bergstra and Y.~Bengio, ``Random search for hyper-parameter optimization,''
  \emph{J. Mach. Learn. Res.}, vol.~13, pp. 281--305, 2012.

\bibitem{snoek12practicalbo}
J.~Snoek, H.~Larochelle, and R.~P. Adams, ``Practical bayesian optimization of
  machine learning algorithms,'' in \emph{Proceedings of the 25th International
  Conference on Neural Information Processing Systems - Volume 2}, ser.
  NIPS'12.\hskip 1em plus 0.5em minus 0.4em\relax Red Hook, NY, USA: Curran
  Associates Inc., 2012, p. 2951–2959.

\bibitem{turner20bovsrs}
R.~Turner, D.~Eriksson, M.~McCourt, J.~Kiili, E.~Laaksonen, Z.~Xu, and
  I.~Guyon, ``Bayesian optimization is superior to random search for machine
  learning hyperparameter tuning: Analysis of the black-box optimization
  challenge 2020,'' in \emph{NeurIPS 2020 Competition and Demonstration Track,
  6-12 December 2020, Virtual Event / Vancouver, BC, Canada}, ser. Proceedings
  of Machine Learning Research, H.~J. Escalante and K.~Hofmann, Eds., vol.
  133.\hskip 1em plus 0.5em minus 0.4em\relax {PMLR}, 2020, pp. 3--26.

\bibitem{hutter_parallel_2012}
F.~Hutter, H.~H. Hoos, and K.~Leyton-Brown, ``Parallel algorithm
  configuration,'' in \emph{Learning and Intelligent Optimization}, ser.
  Lecture Notes in Computer Science, Y.~Hamadi and M.~Schoenauer, Eds.\hskip
  1em plus 0.5em minus 0.4em\relax Berlin, Heidelberg: Springer Berlin
  Heidelberg, 2012, no. 7219, pp. 55--70.

\bibitem{bischl14moimbo}
B.~Bischl, S.~Wessing, N.~Bauer, K.~Friedrichs, and C.~Weihs, ``{MOI-MBO:}
  multiobjective infill for parallel model-based optimization,'' in
  \emph{Learning and Intelligent Optimization - 8th International Conference,
  Lion 8, Gainesville, FL, USA, February 16-21, 2014. Revised Selected Papers},
  ser. Lecture Notes in Computer Science, P.~M. Pardalos, M.~G.~C. Resende,
  C.~Vogiatzis, and J.~L. Walteros, Eds., vol. 8426.\hskip 1em plus 0.5em minus
  0.4em\relax Springer, 2014, pp. 173--186.

\bibitem{gonzalez16batchbo}
J.~Gonz{\'{a}}lez, Z.~Dai, P.~Hennig, and N.~D. Lawrence, ``Batch bayesian
  optimization via local penalization,'' in \emph{Proceedings of the 19th
  International Conference on Artificial Intelligence and Statistics, {AISTATS}
  2016, Cadiz, Spain, May 9-11, 2016}, ser. {JMLR} Workshop and Conference
  Proceedings, A.~Gretton and C.~C. Robert, Eds., vol.~51.\hskip 1em plus 0.5em
  minus 0.4em\relax JMLR.org, 2016, pp. 648--657.

\bibitem{chevalier_fast_2013}
C.~Chevalier and D.~Ginsbourger, ``Fast computation of the multi-points
  expected improvement with applications in batch selection,'' in
  \emph{Learning and Intelligent Optimization}.\hskip 1em plus 0.5em minus
  0.4em\relax Springer Berlin Heidelberg, 2013, pp. 59--69.

\bibitem{balandat2020botorch}
\BIBentryALTinterwordspacing
M.~Balandat, B.~Karrer, D.~R. Jiang, S.~Daulton, B.~Letham, A.~G. Wilson, and
  E.~Bakshy, ``{BoTorch}: A framework for efficient monte-carlo bayesian
  optimization,'' in \emph{Advances in Neural Information Processing Systems},
  H.~Larochelle, M.~Ranzato, R.~Hadsell, M.~F. Balcan, and H.~Lin, Eds.,
  vol.~33, 2020. [Online]. Available:
  \url{https://proceedings.neurips.cc/paper/2020/hash/f5b1b89d98b7286673128a5fb112cb9a-Abstract.html}
\BIBentrySTDinterwordspacing

\bibitem{li2017hyperband}
L.~Li, K.~G. Jamieson, G.~DeSalvo, A.~Rostamizadeh, and A.~Talwalkar,
  ``Hyperband: {A} novel bandit-based approach to hyperparameter
  optimization,'' \emph{J. Mach. Learn. Res.}, vol.~18, pp. 185:1--185:52,
  2017.

\bibitem{li20asha}
L.~Li, K.~G. Jamieson, A.~Rostamizadeh, E.~Gonina, J.~Ben{-}tzur, M.~Hardt,
  B.~Recht, and A.~Talwalkar, ``A system for massively parallel hyperparameter
  tuning,'' in \emph{Proceedings of Machine Learning and Systems 2020, MLSys
  2020, Austin, TX, USA, March 2-4, 2020}, I.~S. Dhillon, D.~S. Papailiopoulos,
  and V.~Sze, Eds.\hskip 1em plus 0.5em minus 0.4em\relax mlsys.org, 2020.

\bibitem{klein20abohb}
L.~C. Tiao, A.~Klein, C.~Archambeau, and M.~W. Seeger, ``Model-based
  asynchronous hyperparameter optimization,'' \emph{CoRR}, vol. abs/2003.10865,
  2020.

\bibitem{falkner18bohb}
S.~Falkner, A.~Klein, and F.~Hutter, ``{BOHB:} robust and efficient
  hyperparameter optimization at scale,'' in \emph{Proceedings of the 35th
  International Conference on Machine Learning, {ICML} 2018,
  Stockholmsm{\"{a}}ssan, Stockholm, Sweden, July 10-15, 2018}, ser.
  Proceedings of Machine Learning Research, J.~G. Dy and A.~Krause, Eds.,
  vol.~80.\hskip 1em plus 0.5em minus 0.4em\relax {PMLR}, 2018, pp. 1436--1445.

\bibitem{tiao20abohb}
L.~C. Tiao, A.~Klein, C.~Archambeau, and M.~W. Seeger, ``Model-based
  asynchronous hyperparameter optimization,'' \emph{CoRR}, vol. abs/2003.10865,
  2020.

\bibitem{sculley15technicaldept}
D.~Sculley, G.~Holt, D.~Golovin, E.~Davydov, T.~Phillips, D.~Ebner,
  V.~Chaudhary, M.~Young, J.~Crespo, and D.~Dennison, ``Hidden technical debt
  in machine learning systems,'' in \emph{Advances in Neural Information
  Processing Systems 28: Annual Conference on Neural Information Processing
  Systems 2015, December 7-12, 2015, Montreal, Quebec, Canada}, C.~Cortes,
  N.~D. Lawrence, D.~D. Lee, M.~Sugiyama, and R.~Garnett, Eds., 2015, pp.
  2503--2511.

\bibitem{hoos_programming_2012}
H.~H. Hoos, ``Programming by {Optimization},'' \emph{Communications of the
  Association for Computing Machinery (CACM)}, vol.~55, no.~2, pp. 70--80, Feb.
  2012.

\bibitem{rice_algorithm_1976}
J.~R. Rice, ``The {Algorithm} {Selection} {Problem},'' \emph{Advances in
  Computers}, vol.~15, pp. 65--118, 1976.

\bibitem{pfahringer_meta-learning_2000}
B.~Pfahringer, H.~Bensusan, and C.~G. Giraud-Carrier, ``Meta-{Learning} by
  {Landmarking} {Various} {Learning} {Algorithms},'' in \emph{Proceedings of
  the {Seventeenth} {International} {Conference} on {Machine} {Learning}}, ser.
  {ICML} ’00.\hskip 1em plus 0.5em minus 0.4em\relax San Francisco, CA, USA:
  Morgan Kaufmann Publishers Inc., 2000, pp. 743--750.

\bibitem{sun_pairwise_2013}
Q.~Sun and B.~Pfahringer, ``\BIBforeignlanguage{English}{Pairwise meta-rules
  for better meta-learning-based algorithm ranking},''
  \emph{\BIBforeignlanguage{English}{Machine Learning}}, vol.~93, no.~1, pp.
  141--161, 2013.

\bibitem{soares_meta-learning_2004}
C.~Soares, P.~B. Brazdil, and P.~Kuba, ``A {Meta}-{Learning} {Method} to
  {Select} the {Kernel} {Width} in {Support} {Vector} {Regression},''
  \emph{Mach. Learn.}, vol.~54, no.~3, pp. 195--209, Mar. 2004.

\bibitem{finn_model-agnostic_2017}
C.~Finn, P.~Abbeel, and S.~Levine, ``Model-{Agnostic} {Meta}-{Learning} for
  {Fast} {Adaptation} of {Deep} {Networks},'' in \emph{Proceedings of the 34th
  {International} {Conference} on {Machine} {Learning}}, ser. Proceedings of
  {Machine} {Learning} {Research}, D.~Precup and Y.~W. Teh, Eds.,
  vol.~70.\hskip 1em plus 0.5em minus 0.4em\relax PMLR, Aug. 2017, pp.
  1126--1135.

\bibitem{vanschoren2018metalearning}
J.~Vanschoren, ``Meta-learning: A survey,'' 2018.

\bibitem{hutter_automated_2019}
F.~Hutter, L.~Kotthoff, and J.~Vanschoren, Eds., \emph{Automated {Machine}
  {Learning}: {Methods}, {Systems}, {Challenges}}, 1st~ed., ser. The {Springer}
  {Series} on {Challenges} in {Machine} {Learning}.\hskip 1em plus 0.5em minus
  0.4em\relax Springer, Cham, 2019.

\bibitem{minton_automatically_1996}
S.~Minton, ``Automatically {Configuring} {Constraint} {Satisfaction}
  {Programs}: {A} {Case} {Study},'' \emph{Constraints}, vol.~1, pp. 7--43,
  1996.

\bibitem{westfold_synthesis_2001}
S.~J. Westfold and D.~R. Smith, ``Synthesis of {Efficient} {Constraint}
  {Satisfaction} {Programs},'' \emph{Knowl. Eng. Rev.}, vol.~16, no.~1, pp.
  69--84, 2001.

\bibitem{balasubramaniam_dominion_2011}
D.~Balasubramaniam, L.~de~Silva, C.~A. Jefferson, L.~Kotthoff, I.~Miguel, and
  P.~Nightingale, ``Dominion: {An} {Architecture}-driven {Approach} to
  {Generating} {Efficient} {Constraint} {Solvers},'' in \emph{9th {Working}
  {IEEE}/{IFIP} {Conference} on {Software} {Architecture}}, Jun. 2011, pp.
  228--231.

\bibitem{monette_aeon_2009}
J.-N. Monette, Y.~Deville, and P.~van Hentenryck, ``Aeon: {Synthesizing}
  {Scheduling} {Algorithms} from {High}-{Level} {Models},'' in \emph{Operations
  {Research} and {Cyber}-{Infrastructure}}, 2009, pp. 43--59.

\bibitem{dorne_hsf_2001}
R.~Dorne and C.~Voudouris, ``{HSF}: {A} {Generic} {Framework} to {Easily}
  {Design} {Meta}-{Heuristic} {Methods},'' in \emph{4th {Metaheuristics}
  {International} {Conference} ({MIC} 2001)}.\hskip 1em plus 0.5em minus
  0.4em\relax John Wiley \& Sons, 2001, pp. 423--428.

\bibitem{ansel_petabricks_2009}
J.~Ansel, C.~Chan, Y.~L. Wong, M.~Olszewski, Q.~Zhao, A.~Edelman, and
  S.~Amarasinghe, ``{PetaBricks}: {A} {Language} and {Compiler} for
  {Algorithmic} {Choice},'' \emph{SIGPLAN Not.}, vol.~44, no.~6, pp. 38--49,
  Jun. 2009.

\bibitem{khudabukhsh_satenstein_2009}
A.~R. KhudaBukhsh, L.~Xu, H.~H. Hoos, and K.~Leyton-Brown, ``{SATenstein}:
  automatically building local search {SAT} solvers from components,'' in
  \emph{Proceedings of the 21st {International} {Joint} {Conference} on
  {Artifical} {Intelligence}}.\hskip 1em plus 0.5em minus 0.4em\relax San
  Francisco, CA, USA: Morgan Kaufmann Publishers Inc., 2009, pp. 517--524.

\bibitem{LopStu2012tec}
M.~L{\'o}pez-Ib{\'a}{\~n}ez and T.~St{\"u}tzle, ``The automatic design of
  multi-objective ant colony optimization algorithms,'' \emph{IEEE Transactions
  on Evolutionary Computation}, vol.~16, no.~6, pp. 861--875, 2012.

\bibitem{lopez2016irace}
M.~L{\'o}pez-Ib{\'a}{\~n}ez, J.~Dubois-Lacoste, L.~P. C{\'a}ceres,
  M.~Birattari, and T.~St{\"u}tzle, ``The irace package: Iterated racing for
  automatic algorithm configuration,'' \emph{Operations Research Perspectives},
  vol.~3, pp. 43--58, 2016.

\bibitem{seipp_new_2015}
J.~Seipp, F.~Pommerening, and M.~Helmert, ``New {Optimization} {Functions} for
  {Potential} {Heuristics},'' in \emph{Twenty-{Fifth} {International}
  {Conference} on {International} {Conference} on {Automated} {Planning} and
  {Scheduling}}, ser. {ICAPS}'15.\hskip 1em plus 0.5em minus 0.4em\relax AAAI
  Press, 2015, pp. 193--201.

\bibitem{malkomes2018automating}
G.~Malkomes and R.~Garnett, ``Automating bayesian optimization with bayesian
  optimization,'' \emph{Advances in Neural Information Processing Systems},
  vol.~31, pp. 5984--5994, 2018.

\bibitem{lindauer2019assessing}
M.~Lindauer, M.~Feurer, K.~Eggensperger, A.~Biedenkapp, and F.~Hutter,
  ``Towards assessing the impact of bayesian optimization's own
  hyperparameters,'' 2019.

\bibitem{metairace}
N.~Dang, P.~{De Causmaecker}, L.~C{\'a}ceres, and T.~St{\"u}tzle,
  ``\BIBforeignlanguage{English}{Configuring irace using surrogate
  configuration benchmarks},'' in \emph{\BIBforeignlanguage{English}{GECCO 2017
  - Proceedings of the 2017 Genetic and Evolutionary Computation Conference}},
  ser. GECCO 2017 - Proceedings of the 2017 Genetic and Evolutionary
  Computation Conference.\hskip 1em plus 0.5em minus 0.4em\relax Association
  for Computing Machinery, Inc, Jul. 2017, pp. 243--250, publisher Copyright:
  {\textcopyright} 2017 ACM.; null ; Conference date: 15-07-2017 Through
  19-07-2017.

\bibitem{fawcett2016analysisAblation}
C.~Fawcett and H.~H. Hoos, ``Analysing differences between algorithm
  configurations through ablation,'' \emph{J. Heuristics}, vol.~22, no.~4, pp.
  431--458, 2016.

\bibitem{moaco}
M.~López-Ibáñez and T.~Stützle, ``An experimental analysis of design
  choices of multi-objective ant colony optimization algorithms,'' \emph{Swarm
  Intelligence}, vol.~6, pp. 207--232, 2012.

\bibitem{bischl_evco12a}
B.~Bischl, O.~Mersmann, H.~Trautmann, and C.~Weihs, ``Resampling methods for
  meta-model validation with recommendations for evolutionary computation,''
  \emph{Evolutionary Computation}, vol.~20, no.~2, pp. 249--275, 2012.

\bibitem{klein2020model}
A.~Klein, L.~C. Tiao, T.~Lienart, C.~Archambeau, and M.~Seeger, ``Model-based
  asynchronous hyperparameter and neural architecture search,'' \emph{arXiv
  preprint arXiv:2003.10865}, 2020.

\bibitem{klein2017fast}
A.~Klein, S.~Falkner, S.~Bartels, P.~Hennig, and F.~Hutter, ``Fast bayesian
  optimization of machine learning hyperparameters on large datasets,'' in
  \emph{Artificial Intelligence and Statistics}.\hskip 1em plus 0.5em minus
  0.4em\relax PMLR, 2017, pp. 528--536.

\bibitem{jamieson16successivehalving}
K.~G. Jamieson and A.~Talwalkar, ``Non-stochastic best arm identification and
  hyperparameter optimization,'' in \emph{Proceedings of the 19th International
  Conference on Artificial Intelligence and Statistics, {AISTATS} 2016, Cadiz,
  Spain, May 9-11, 2016}, ser. {JMLR} Workshop and Conference Proceedings,
  A.~Gretton and C.~C. Robert, Eds., vol.~51.\hskip 1em plus 0.5em minus
  0.4em\relax JMLR.org, 2016, pp. 240--248.

\bibitem{bergstra11treeparzen}
J.~Bergstra, R.~Bardenet, Y.~Bengio, and B.~K{\'e}gl, ``Algorithms for
  hyper-parameter optimization,'' \emph{Advances in neural information
  processing systems}, vol.~24, 2011.

\bibitem{golovin2017google}
D.~Golovin, B.~Solnik, S.~Moitra, G.~Kochanski, J.~Karro, and D.~Sculley,
  ``Google vizier: A service for black-box optimization,'' in \emph{Proceedings
  of the 23rd ACM SIGKDD international conference on knowledge discovery and
  data mining}, 2017, pp. 1487--1495.

\bibitem{birattari2009tuning}
M.~Birattari and J.~Kacprzyk, \emph{Tuning metaheuristics: a machine learning
  perspective}.\hskip 1em plus 0.5em minus 0.4em\relax Springer, 2009, vol.
  197.

\bibitem{birattari2010f}
M.~Birattari, Z.~Yuan, P.~Balaprakash, and T.~St{\"u}tzle, ``F-race and
  iterated f-race: An overview,'' \emph{Experimental methods for the analysis
  of optimization algorithms}, pp. 311--336, 2010.

\bibitem{hutter2009paramils}
F.~Hutter, H.~H. Hoos, K.~Leyton-Brown, and T.~St{\"u}tzle, ``Paramils: an
  automatic algorithm configuration framework,'' \emph{Journal of Artificial
  Intelligence Research}, vol.~36, pp. 267--306, 2009.

\bibitem{ansotegui2009gender}
C.~Ans{\'o}tegui, M.~Sellmann, and K.~Tierney, ``A gender-based genetic
  algorithm for the automatic configuration of algorithms,'' in
  \emph{International Conference on Principles and Practice of Constraint
  Programming}.\hskip 1em plus 0.5em minus 0.4em\relax Springer, 2009, pp.
  142--157.

\bibitem{maron1997racing}
O.~Maron and A.~W. Moore, ``The racing algorithm: Model selection for lazy
  learners,'' in \emph{Lazy learning}.\hskip 1em plus 0.5em minus 0.4em\relax
  Springer, 1997, pp. 193--225.

\bibitem{saltelli02saimportance}
A.~Saltelli, ``Sensitivity analysis for importance assessment,'' \emph{Risk
  Analysis}, vol.~22, no.~3, pp. 579--590, 2002.

\bibitem{hutter14fanova}
F.~Hutter, H.~H. Hoos, and K.~Leyton{-}Brown, ``An efficient approach for
  assessing hyperparameter importance,'' in \emph{Proceedings of the 31th
  International Conference on Machine Learning, {ICML} 2014, Beijing, China,
  21-26 June 2014}, ser. {JMLR} Workshop and Conference Proceedings,
  vol.~32.\hskip 1em plus 0.5em minus 0.4em\relax JMLR.org, 2014, pp. 754--762.

\bibitem{cheng2020surrogate}
K.~Cheng, Z.~Lu, C.~Ling, and S.~Zhou, ``Surrogate-assisted global sensitivity
  analysis: an overview,'' \emph{Structural and Multidisciplinary
  Optimization}, vol.~61, no.~3, pp. 1187--1213, 2020.

\bibitem{sheikholeslami21autoablation}
S.~Sheikholeslami, M.~Meister, T.~Wang, A.~H. Payberah, V.~Vlassov, and
  J.~Dowling, ``Autoablation: Automated parallel ablation studies for deep
  learning,'' in \emph{EuroMLSys@EuroSys 2021, Proceedings of the 1st Workshop
  on Machine Learning and Systemsg Virtual Event, Edinburgh, Scotland, UK, 26
  April, 2021}, E.~Yoneki and P.~Patras, Eds.\hskip 1em plus 0.5em minus
  0.4em\relax {ACM}, 2021, pp. 55--61.

\bibitem{pfisterer_yahpo_2021}
F.~Pfisterer, L.~Schneider, J.~Moosbauer, M.~Binder, and B.~Bischl, ``{YAHPO}
  {Gym} -- {Design} {Criteria} and a new {Multifidelity} {Benchmark} for
  {Hyperparameter} {Optimization},'' \emph{arXiv:2109.03670 [cs, stat]}, 2021,
  arXiv: 2109.03670.

\bibitem{cheng2016wide}
H.-T. Cheng, L.~Koc, J.~Harmsen, T.~Shaked, T.~Chandra, H.~Aradhye,
  G.~Anderson, G.~Corrado, W.~Chai, M.~Ispir \emph{et~al.}, ``Wide \& deep
  learning for recommender systems,'' in \emph{Proceedings of the 1st workshop
  on deep learning for recommender systems}, 2016, pp. 7--10.

\bibitem{ZimLin2021a}
L.~Zimmer, M.~Lindauer, and F.~Hutter, ``Auto-pytorch tabular: Multi-fidelity
  metalearning for efficient and robust autodl,'' \emph{IEEE Transactions on
  Pattern Analysis and Machine Intelligence}, vol.~43, no.~9, pp. 3079 -- 3090,
  2021.

\bibitem{siems2020nasbench301}
J.~Siems, L.~Zimmer, A.~Zela, J.~Lukasik, M.~Keuper, and F.~Hutter,
  ``Nas-bench-301 and the case for surrogate benchmarks for neural architecture
  search,'' 2020.

\bibitem{vanschoren2013openML}
J.~Vanschoren, J.~N. van Rijn, B.~Bischl, and L.~Torgo, ``Openml: networked
  science in machine learning,'' \emph{{SIGKDD} Explor.}, vol.~15, no.~2, pp.
  49--60, 2013.

\bibitem{hnsw}
Y.~A. Malkov and D.~A. Yashunin, ``Efficient and robust approximate nearest
  neighbor search using hierarchical navigable small world graphs,'' \emph{IEEE
  transactions on pattern analysis and machine intelligence}, vol.~42, no.~4,
  pp. 824--836, 2018.

\bibitem{friedman10glmnet}
J.~Friedman, T.~Hastie, and R.~Tibshirani, ``Regularization paths for
  generalized linear models via coordinate descent,'' \emph{Journal of
  Statistical Software}, vol.~33, no.~1, pp. 1--22, 2010.

\bibitem{wright17ranger}
M.~N. Wright and A.~Ziegler, ``{ranger}: A fast implementation of random
  forests for high dimensional data in {C++} and {R},'' \emph{Journal of
  Statistical Software}, vol.~77, no.~1, pp. 1--17, 2017.

\bibitem{rpart}
L.~Breiman, J.~H. Friedman, R.~A. Olshen, and C.~J. Stone, \emph{Classification
  And Regression Trees}.\hskip 1em plus 0.5em minus 0.4em\relax Routledge, Oct.
  2017.

\bibitem{boser92svm}
B.~E. Boser, I.~Guyon, and V.~Vapnik, ``A training algorithm for optimal margin
  classifiers,'' in \emph{Proceedings of the Fifth Annual {ACM} Conference on
  Computational Learning Theory, {COLT} 1992, Pittsburgh, PA, USA, July 27-29,
  1992}, D.~Haussler, Ed.\hskip 1em plus 0.5em minus 0.4em\relax {ACM}, 1992,
  pp. 144--152.

\bibitem{chen16xgboost}
T.~Chen and C.~Guestrin, ``{XGBoost}: A scalable tree boosting system,'' in
  \emph{Proceedings of the 22nd ACM SIGKDD International Conference on
  Knowledge Discovery and Data Mining}, ser. KDD '16.\hskip 1em plus 0.5em
  minus 0.4em\relax New York, NY, USA: ACM, 2016, pp. 785--794.

\bibitem{binder2020}
M.~Binder, F.~Pfisterer, and B.~Bischl, ``Collecting empirical data about
  hyperparameters for data driven automl,'' in \emph{Proceedings of the 7th
  ICML Workshop on Automated Machine Learning (AutoML 2020)}, 2020.

\bibitem{liu19}
H.~Liu, K.~Simonyan, and Y.~Yang, ``{DARTS}: {Differentiable} {Architecture}
  {Search},'' in \emph{Proceedings of the International Conference on Learning
  Representations}, 2019.

\bibitem{cifar}
A.~Krizhevsky, ``Learning multiple layers of features from tiny images,''
  University of Toronto, Tech. Rep., 2009.

\bibitem{breimanrandomforest}
L.~Breiman, ``Random forests,'' \emph{Mach. Learn.}, vol.~45, no.~1, pp. 5--32,
  2001.

\bibitem{Samworth2012}
R.~J. Samworth, ``Optimal weighted nearest neighbour classifiers,'' \emph{The
  Annals of Statistics}, vol.~40, no.~5, Oct. 2012.

\bibitem{demsar06cd}
J.~Demsar, ``Statistical comparisons of classifiers over multiple data sets,''
  \emph{J. Mach. Learn. Res.}, vol.~7, pp. 1--30, 2006.

\bibitem{mersmann2011exploratory}
O.~Mersmann, B.~Bischl, H.~Trautmann, M.~Preuss, C.~Weihs, and G.~Rudolph,
  ``Exploratory landscape analysis,'' in \emph{Proceedings of the 13th annual
  conference on Genetic and evolutionary computation}, 2011, pp. 829--836.

\end{thebibliography}
